\documentclass{bmvc2k}

\usepackage{times}
\usepackage{epsfig}
\usepackage{graphicx}
\usepackage{amsmath}
\usepackage{amssymb}
\usepackage{booktabs}\usepackage{algorithm}
\usepackage{algpseudocode}
\usepackage{graphicx, array, blindtext}

\usepackage{multirow}
\usepackage[symbol]{footmisc}

\usepackage{array, booktabs, makecell}
\usepackage{siunitx, mhchem}
\usepackage{enumitem}\newcommand{\MainFramework}{Self-Supervised Robustifying Guidance}
\newcommand{\MainFrameworkBold}{Self-Supervised \textbf{RO}bustifying \textbf{GU}idanc\textbf{E}}
\newcommand{\MainShort}{ROGUE}
\newcommand{\RobustPipeline}{Robustification Pipeline}
\newcommand{\GuidePipeline}{Guidance Pipeline}
\newcommand{\noiserobust}{noise-robustification}
\newcommand{\occlusionrobust}{occlusion-robustification}

\newcommand{\Synthetic}{SynChOcc}

\newcommand{\Real}{ReaChOcc}
\title{Self-Supervised Robustifying Guidance for Monocular 3D Face Reconstruction}

\addauthor{Hitika Tiwari}{hitika@iitk.ac.in}{1,4,$\ast$}
\addauthor{Min-Hung Chen}{vitec6@gmail.com}{2,$\ast\ast$}
\addauthor{Yi-Min Tsai}{yi-min.tsai@mediatek.com}{3}
\addauthor{Hsien-Kai Kuo}{hsienkai.kuo@mediatek.com}{3}
\addauthor{Hung-Jen Chen}{hj.chen@mediatek.com}{3}
\addauthor{Kevin Jou}{kevin.jou@mediatek.com}{3}
\addauthor{K. S. Venkatesh}{venkats@iitk.ac.in}{1}
\addauthor{Yong-Sheng Chen}{yschen@nycu.edu.tw}{4}
\addinstitution{Indian Institute of Technology Kanpur, India
}
\addinstitution{{Microsoft AI R\&D Center, Taiwan}
}
\addinstitution{{MediaTek Inc., Taiwan}
}
\addinstitution{National Yang Ming Chiao Tung University, Taiwan
}

\runninghead{Tiwari et al.}{{Self-Supervised Robust Monocular Face Reconstruction}}

\begin{document}

\maketitle
\begin{abstract}
   Despite the recent developments in 3D Face Reconstruction from occluded and noisy face images, the performance is still unsatisfactory. 
   Moreover, most existing methods rely on additional dependencies, posing numerous constraints over the training procedure.  
   Therefore, we propose a \textit{\MainFrameworkBold~(\textbf{\MainShort})} framework to obtain robustness against occlusions and noise in the face images. The proposed network contains 1) the \textit{\GuidePipeline}~to obtain the $3$D face coefficients for the clean faces and 2) the \textit{\RobustPipeline}~to acquire the consistency between the estimated coefficients for occluded or noisy images and the clean counterpart. The proposed image- and feature-level loss functions aid the \MainShort~learning process without posing additional dependencies. To facilitate model evaluation, we propose two challenging occlusion face datasets, \textbf{\textit{ReaChOcc}} and \textbf{SynChOcc}, containing real-world and synthetic occlusion-based face images for robustness evaluation. Also, a noisy variant of the test dataset of CelebA is produced for evaluation. Our method outperforms the current state-of-the-art method by large margins
   (e.g., for the perceptual errors, a reduction of $\mathbf{23.8\%}$ for real-world occlusions, $\mathbf{26.4\%}$ for synthetic occlusions, and $\mathbf{22.7\%}$ for noisy images),
\footnotetext{\hspace{-2.1em}$^\ast$Work partially done during the research internship at MediaTek Inc., Taiwan.}
\footnotetext{\hspace{-2.1em}$^\ast$$^\ast$Work partially done during MediaTek Inc., Taiwan.}
   demonstrating the effectiveness of the proposed approach. The occlusion datasets and the corresponding evaluation code are released publicly at \href{https://github.com/ArcTrinity9/Datasets-ReaChOcc-and-SynChOcc}{https://github.com/ArcTrinity9/Datasets-ReaChOcc-and-SynChOcc}.

\end{abstract}


\section{Introduction}
\label{sec:intro}

\begin{figure}[!ht]
\centering
    \includegraphics[width=0.95\textwidth]{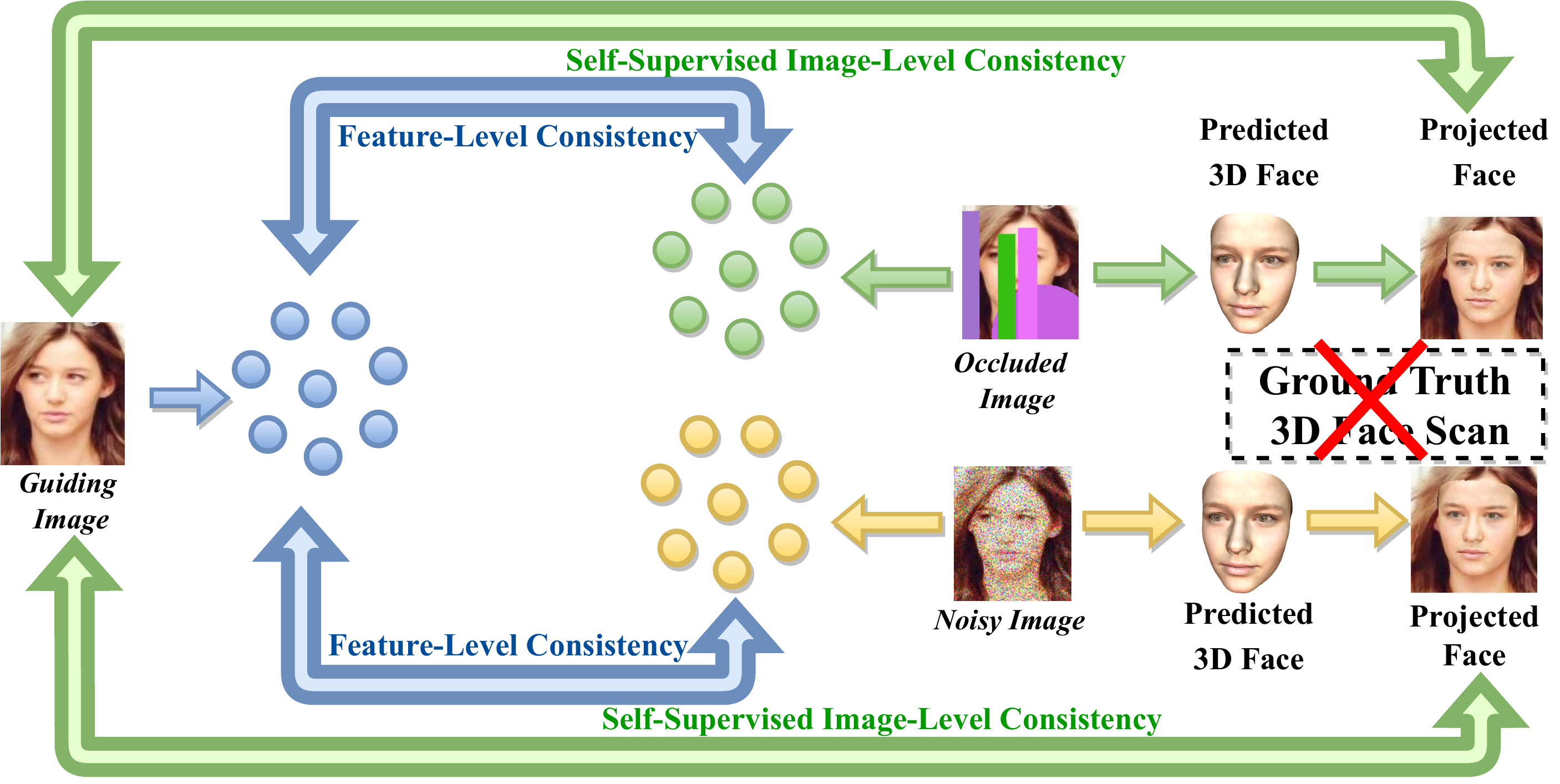}
    \caption{
    An overview of the proposed \textit{\MainFrameworkBold~(\textbf{\MainShort})} framework. \MainShort~addresses the occlusion and noise problems in face images for $3$D face reconstruction by the proposed novel image- and feature-level consistency loss functions in the self-supervised fashion, enforcing occluded and noise coefficients to be consistent with the target coefficients of the guiding image, without the requirement of $3$D ground-truth face scans and any additional dependency for the training. }
    \label{fig:teaser}
\end{figure}


$3$D face reconstruction from monocular face images has been a longstanding problem in the field of $3$D computer graphics and computer vision. Recent deep-learning-based approaches demonstrate encouraging progress
with regard to perceptual accuracy and training efficiency, facilitating
numerous applications such as face recognition~\cite{blanz2003face,tuan2017regressing,pears20203d,adjabi2020past}, face artifice and animation~\cite{jian2021realistic,ye20203d,deng2021plausible}. 
To address the mathematically ill-posed issue, the fitting-based method, \textit{3D Morphable Model ($3$DMM)}~\cite{blanz1999morphable}, proposes a low-dimensional 3DMM search space spanning the range of human facial appearance. The coordinates from the two sub-spaces, \textit{geometry} and \textit{texture}, along with the illumination and pose parameters, generate a $3$D face such that the corresponding face image (projection of $3$D face) resembles the target image. 
However, most target images contain occlusions such as glasses and masks. Moreover, face images are usually not noise-free. Therefore, the fitting-based methods may drift the coordinates outside the $3$DMM space or distort the $3$D face geometry and texture, posing challenges to the problem of $3$D face reconstruction from monocular images. \\
To address the above issues, several approaches have been proposed. Fitting-based optimization approaches~\cite{egger2016occlusion} iteratively adapt the segmentation map to the target face image. $3$D faces can also be obtained from occluded face images using training methodologies with different supervisions~\cite{tewari2018high,tran2018extreme,genova2018unsupervised,deng2019accurate,yuan2019face}.
In addition, depth-based methods~\cite{zhong2020face,li2021robust} tackle noise issues for $3$D face reconstruction with depth maps.
However, the above methods hold several dependencies, such as skin masks, depth maps, ground-truth data, synthetic data, segmented maps, multi-images, etc., posing numerous constraints over the training procedure.  Therefore, a novel training pipeline that can avoid the above-stated requisites and attain robustness against facial occlusions and image noise is desired.  Moreover, there is a need for dedicated occlusion datasets to facilitate the performance evaluation of such models.\\
In this work, we propose two natural occlusion-based test datasets: \textbf{Rea}l World
\textbf{Ch}allenging \textbf{Occ}lusion (\textbf{\Real}), and \textbf{Syn}thetic \textbf{Ch}allenging \textbf{Occ}lusion (\textbf{\Synthetic}) datasets to facilitate robust face reconstruction research, which is not well-explored in the community. Also, we propose a novel \textbf{\textit{\MainFrameworkBold~(\textbf{\MainShort})}} framework, which learns statistical facial coefficients for occluded, and noisy face images simultaneously in a self-supervised manner, without requiring ground truth $3$D face scans. 
The proposed \MainShort~contains two parts: 1) The \textit{\GuidePipeline} estimates coefficients for the clean target face using self-supervised cycle-consistent manners, and 2) the \textit{\RobustPipeline} enforces the estimated coefficients of occluded and noisy faces to be consistent with clean images. The training is done without additional dependencies due to our image and feature-level losses. 
The proposed \MainShort~framework is evaluated on three datasets: \Real, \Synthetic, noise variant of the CelebA~\cite{liu2015deep} dataset, and outperforms the current state-of-the-art methods by large margins. 
For example, for the \textit{perceptual error}, \MainShort~achieves a reduction of $\mathbf{23.8\%}$ ($1.237 \rightarrow 0.943$) for real-world occlusions, $\mathbf{26.4\%}$ ($1.195 \rightarrow 0.879$) for synthetic occlusions, and $\mathbf{22.7\%}$ ($1.245 \rightarrow 0.963$) for noisy images.\\
In summary, the contributions of our work are as follows:
\begin{enumerate}[leftmargin=*]
    \item \textbf{\Real~and~\Synthetic~Testing Datasets}: To facilitate robust face reconstruction research, we propose \Real~and \Synthetic~datasets containing natural real-world and synthetic facial occlusions. Our datasets facilitate both \textit{shape} and \textit{texture} comparisons. We have publicly released the datasets and the corresponding evaluation code.
    
    
    
    \item \textbf{\MainFramework~Framework}: 
    We propose a self-supervised framework with novel image- and feature-level robustification losses, dubbed \textit{\textbf{\MainShort}}, to obtain accurate $3$D faces by attaining robustness against the challenging facial \textit{occlusions} and \textit{noise} in the facial images (e.g., \textit{25+\%} perceptual error reduction), without posing dependencies and the requirement of $3$D ground truth.
\end{enumerate}

\section{Related Work}\label{sec:RelatedWork}
\noindent\textbf{Robustness for Face Reconstruction}: 
Egger et al.~\cite{egger2016occlusion} aim to address the occlusion issues by segmenting the target image into face and non-face regions and iteratively adapting the face model and the segmentation to the target image. Tran et al.~\cite{tran2018extreme} deploy an example-based hole ﬁlling approach by utilizing the reference set of images containing a suitably similar individual as in the target image.
Genova et al.~\cite{genova2018unsupervised} exploit synthetic ground truth data (with the label-free instances of real target image) to tackle the occlusions. Yuan et al.~\cite{yuan2019face} exploit $3$DMM to tackle the occlusions in $2$D images, where the $3$D ground truth data obtained by 3DDFA~\cite{zhu2016face} is required. However, the above methods either only well tackle small-scale occlusions (e.g., minor beards, goggles) instead of large-scale ones (e.g., face masks, tattoos)~\cite{egger2016occlusion,deng2019accurate} or rely upon additional dependencies, such as additional images, synthetic data, $3$D ground truth, etc.~\cite{tran2018extreme,genova2018unsupervised,deng2019accurate,yuan2019face}. Besides, our method focuses on tackling large-scale occlusions without posing additional dependencies. Moreover, the noise in the face images poses a challenge in obtaining accurate $3$D faces. 
To our knowledge, there are no $3$D face reconstruction methods~\cite{egger2016occlusion,tewari2017mofa,tewari2018self,tewari2020stylerig,deng2019accurate,deng2020disentangled,tran2018extreme,tran2018nonlinear,genova2018unsupervised,lin2020towards,chen2019photo,gecer2019ganfit} aiming to reconstruct the $3$D faces from the heavily noisy face images. However, there are depth-based methods~\cite{zhong2020face,li2021robust} aiming to address the issues of device-specific noise in obtaining the depth map for reconstructing $3$D faces, but tackling the noises in the face images is beyond the scope of those papers. In this paper, the proposed \textit{\MainFramework~framework} aims to attain robustness against the image noise and facial occlusions, thus facilitating the accurate reconstruction of $3$D faces from noisy and occluded images.

\begin{figure*}[!t]
\centering
    \includegraphics[width=\textwidth]{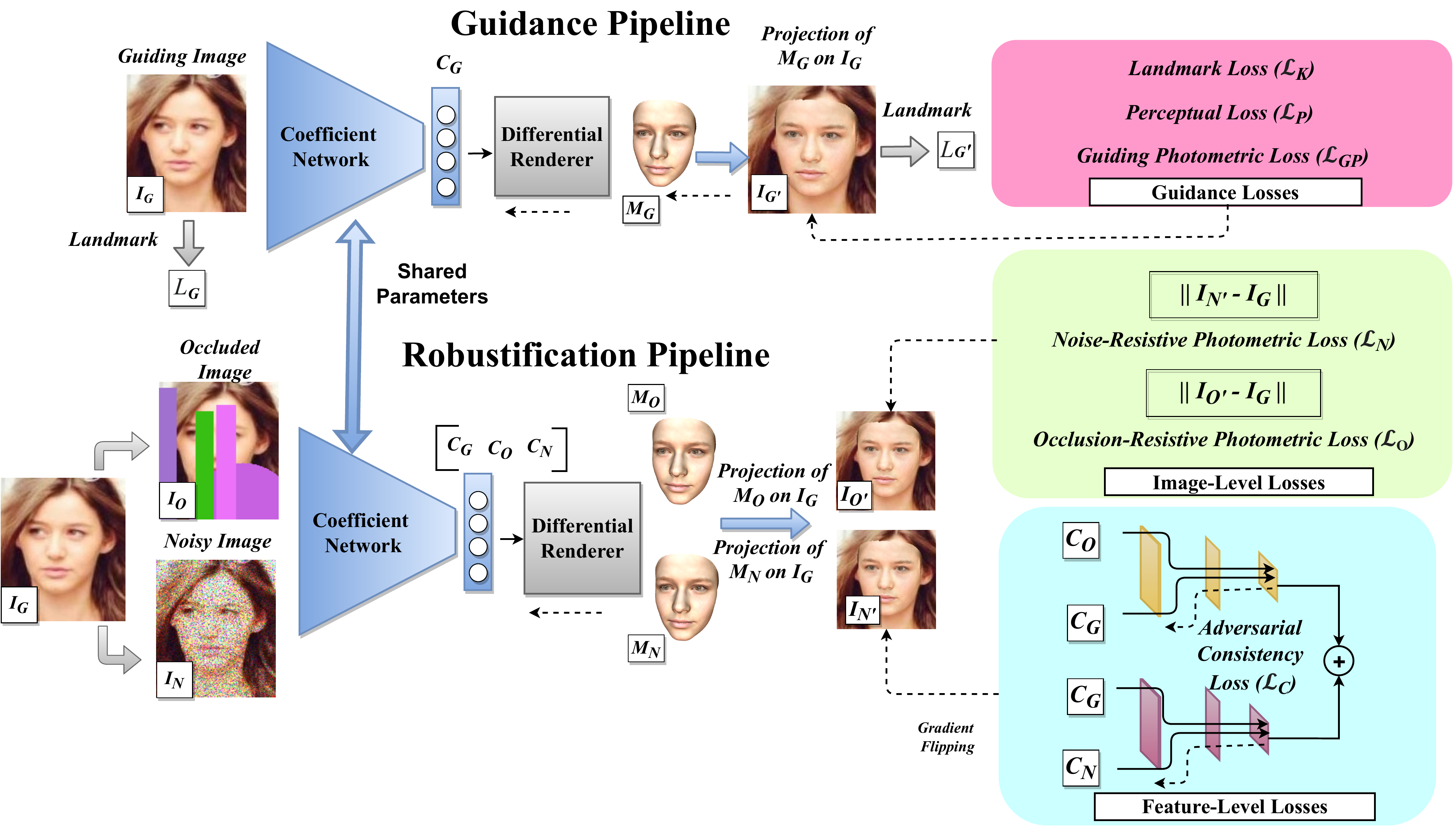}
    \caption{The overall training pipeline of the proposed \textit{\MainFrameworkBold~(\textbf{\MainShort})} framework.
    The \textit{\GuidePipeline} ensures the faithful reconstruction of the $3$D faces from clean \textit{guiding} images $I_G$ in a cycle-consistent manner, and the \textit{\RobustPipeline} enforces the estimated coefficients of occluded and noisy images ($C_O$, $C_N$) to be consistent with \textit{guiding} ones ($C_G$). The training is done in a self-supervised fashion by the proposed self-supervised image-level losses ($\mathcal{L}_{O}$, $\mathcal{L}_{N}$) and feature-level adversarial consistency loss ($\mathcal{L}_{C}$), without the need for $3$D ground truth. Here the solid lines represent the data flow, whereas the dotted lines indicate gradient flow.}
    \label{fig:training}
\end{figure*}

\noindent\textbf{Occlusion-Aware Datasets}: Existing real-world and synthetically occluded test datasets~\cite{RingNet:CVPR:2019, voo2022delving, nojavanasghari2017hand2face,lee2020maskgan} have the following shortcomings: lack of non-occluded ground-truth face images, restrictions on open access, and the limited number of facial occlusions. RealOcc-Wild dataset~\cite{voo2022delving} contains $270$ faces with various natural occlusions. The real-world dataset in~\cite{lee2020maskgan} consists of challenging occlusions (e.g., sunglasses, food, hats, and hands). \cite{nojavanasghari2017hand2face} contains hand-occluded face images. NoW dataset~\cite{RingNet:CVPR:2019} consists of $528$ images with common occlusions. However, these datasets do not contain textured $2$D/$3$D ground-truth data, and some occlusions such as bangs, beards, mustaches, turbans, and masks are absent. In addition, the unavailability of test datasets~\cite{RingNet:CVPR:2019} in the open public domain poses constraints over the testing. Besides, our  publicly released datasets are designed explicitly for reconstruction tasks and contain various occluded images and the corresponding non-occluded faces.



\section{\MainFrameworkBold~(\textbf{\MainShort})}

Despite the encouraging results obtained by the previous methods for $3$D face reconstruction from occluded face images, there is still a large room for improvement with regards to moderately to heavily occluded face images. 
In addition, tackling image noise is still an under-addressed issue. 
Moreover, these methods require several dependencies such as synthetic data, skin masks, etc., posing constraints for training (see Sec.~\ref{sec:RelatedWork} for more details). 
Therefore, we aim to learn $3$D faces in a self-supervised manner without requiring ground truth $3$D face scans and other dependencies. To achieve this goal, we propose the \textbf{\textit{\MainFrameworkBold~(\textbf{\MainShort})}} framework, which is composed of: 1) the \textbf{\textit{\GuidePipeline}} and 2) the \textbf{\textit{\RobustPipeline}} (Fig.~\ref{fig:training}).
For the preliminaries of monocular $3$D face reconstruction, please refer to the Supplementary.

\noindent\textbf{\GuidePipeline}: 
In occlusion robust monocular $3$D face reconstruction, one of the main goals is learning reliable $3$DMM coefficients with \textit{the least} supervision and dependencies. Inspired by R-Net~\cite{deng2019accurate} which contains comparatively fewer dependencies, we propose the \textit{Self-Supervised \GuidePipeline} to learn the coefficients $\boldsymbol{C_G}$ by exploiting the cycle-consistency in a self-supervised manner, as shown in Fig.~\ref{fig:training} (upper).
More specifically, the \textit{\GuidePipeline}~takes a clean (i.e., non-occluded noise-free) image $I_G$ (named \textit{guiding image}) as the input, renders the $3$D mesh $M_G$, and projects back to get the $2$D face image $I_{G'}$. And then $\boldsymbol{C_G}$ is learned by enforcing the consistency between $I_G$ and $I_{G'}$, using only a single monocular face image. Moreover, $\boldsymbol{C_G}$ guides the \RobustPipeline~to attain robustness against the face occlusions and noise in the images \textit{without relying upon external guidance} such as skin masks~\cite{deng2019accurate}, synthetic data~\cite{genova2018unsupervised}, etc. For more details on various components of the \textit{\GuidePipeline} please refer to the Supplementary.

\noindent\textbf{\RobustPipeline}:
Although the \GuidePipeline~reduces the requirement of supervision and dependencies, the two significant issues for monocular $3$D face reconstruction are still not fully addressed: \textit{occlusion} and \textit{noise}.
First of all, current methods still cannot reasonably handle the face images with the majority of facial regions occluded, where these methods drift away from their searches from the $3$DMM space, resulting in the reconstruction of non-human-like $3$D faces. Moreover, additional dependencies such as pre-trained face segmentation models~\cite{lin2020towards}, skin masks~\cite{deng2019accurate}, etc., used by existing methods for tackling the occlusion issues constrain the efficiency of training. 
Furthermore, despite the progress in the $3$D face reconstruction field, no approach has been proposed to tackle the issue of noise in the face image. 
All the above challenges motivate the need to learn $3$D facial coefficients from occluded and noisy face images more accurately and efficiently. 
Therefore, we propose the \textit{Self-Supervised \RobustPipeline}~to attain robustness against the occlusions and noise in the face images \textit{with the least additional dependencies}, as shown in Fig.~\ref{fig:training} (lower).
More specifically, we exploit the guiding image $I_G$ and the estimated coefficients $\boldsymbol{C_G}$ from \GuidePipeline, and encourage the geometry and texture consistency between the \RobustPipeline~and the \GuidePipeline, to make $\boldsymbol{C_G}$ consistent with the estimated coefficients $\boldsymbol{C_O}$ (from occluded face images $I_O$) and $\boldsymbol{C_N}$ (from noisy face images $I_N$). All the components of the \RobustPipeline~are presented as follows: \\
\noindent 1) To \textbf{obtain consistency with the \GuidePipeline} for the Robustification (occlusion and noise) coefficients, 
we exploit a three-layer 
Generative Adversarial Network (GAN) architecture and propose the \textit{Adversarial Consistency Loss} $\mathcal{L}_C$ as follows:
\begin{align}\label{eq:c}
    \mathcal{L}_C = \mathcal{L}_{CO}+\mathcal{L}_{CN}, 
   \mathcal{L}_{CO} = \mathcal{L}_{h}(\mathcal{D}(\boldsymbol{C_G},\boldsymbol{C_O}), [d_G, d_O]),
   \mathcal{L}_{CN} = \mathcal{L}_{h}(\mathcal{D}(\boldsymbol{C_G},\boldsymbol{C_N}), [d_G, d_N]),
\end{align}
where $\mathcal{L}_{CO}$ represents the \textit{\occlusionrobust~consistency loss} for tackling the occlusion issues and $\mathcal{L}_{CN}$ denotes \textit{\noiserobust~consistency loss} for tackling the noise in the face image.
In the equation, $\mathcal{D}$ is the classifier to discriminate $\boldsymbol{C_G}$ and $\boldsymbol{C_i}\in\mathbb{R}^{257}$ ($i=O/N$), and $\mathcal{L}_h$ denotes the standard Huber loss function. In addition, $d_i\in\mathbb{R}$ ($i = G/O/N$) represents the labels associated with the (guiding/occlusion/noise) coefficients. \\
\noindent 2) To \textbf{ensure the guidance direction}
such that the \RobustPipeline~learns through the experience of the \GuidePipeline~and not vice-versa, we directly regress the pixels of the projected $3$D face obtained from the occluded face images ($I_{O'}$) and noisy face images ($I_{N'}$) over the guidance counterpart ($I_G$) by the proposed \textit{Occlusion-Resistive Photometric Loss} $\mathcal{L}_{O}$ and \textit{Noise-Resistive Photometric Loss} $\mathcal{L}_{N}$, respectively, as follows:
\begin{align}\label{eq:i}
    \mathcal{L}_{O}= || I_{O'}-I_G ||,\quad \mathcal{L}_{N} = || I_{N'}-I_G ||.
\end{align}


The overall loss function $\mathcal{L}_{robust}$ for the proposed method
can be expressed below:
\begin{align}\label{EQ:follow}
    \mathcal{L}_{robust} = \beta_O\mathcal{L}_{O}+\beta_N\mathcal{L}_{N} - \beta_C\mathcal{L}_{C},
\end{align}
where $\beta_O,\beta_N$ and $\beta_C$ are the weights associated with occlusion and noise-resistive photometric losses (Eq.~\eqref{eq:i}), and adversarial consistency loss (Eq.~\eqref{eq:c}), respectively. The negative sign indicates the \textit{adversarial training}. For simplicity, the notation of the image index is ignored here. It is worth noting that the proposed \textit{\MainFramework} framework leverages the novel robustification loss function $\mathcal{L}_{robust}$. Thus our approach bears a significant difference from R-Net~\cite{deng2019accurate} regarding the model, architecture, losses, and target data. Unlike R-Net, our model does not require skin masks for the training, facilitating training efficiency. Moreover, our proposed framework is the first (to the best of our knowledge) to tackle the noise in the face images for $3$D face reconstruction without $3$D ground truth.

\section{Dataset Preparation}\label{sec:dataset}
\begin{figure}[ht]
    \centering
    \includegraphics[width=0.9\textwidth]{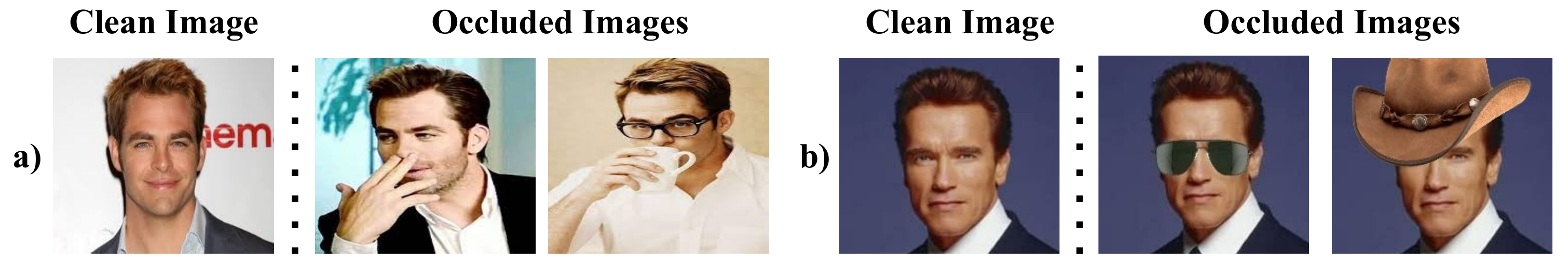}
    \caption{A demonstration of the samples from the proposed a) \Real, and b) \Synthetic~datasets. Our \Real~contains \textit{unpaired clean images}, whereas \Synthetic~provides \textit{paired clean images} for comparison.}
    \label{fig:data}
\end{figure}

To obtain the training data, we exploit the training set of several standard face datasets as the clean \textit{guiding} images and create synthetic \textit{occluded} and \textit{noisy} face images for our training pipeline. For testing, numerous real-world and synthetically occluded test datasets~\cite{RingNet:CVPR:2019, voo2022delving, nojavanasghari2017hand2face,lee2020maskgan} have been proposed. However, these datasets have shortcomings such as the unavailability of the dataset in the public domain, lack of non-occluded ground-truth face images, and a limited number of facial occlusions. RealOcc-Wild dataset~\cite{voo2022delving} contains $270$ faces with various natural occlusions. The real-world dataset in~\cite{lee2020maskgan} consists of challenging occlusions (e.g., sunglasses, food, hats, and hands). \cite{nojavanasghari2017hand2face} contains hand-occluded face images. For $3$D face reconstruction, there is only one occlusion-based dataset, NoW test set~\cite{RingNet:CVPR:2019}, which is not publicly available and thus poses constraints on the testing. Moreover, the datasets mentioned above do not contain several types of facial occlusions such as bangs, beards, mustaches, turbans, and masks. 
Therefore, a dataset is required which contains numerous possible occlusions and the corresponding non-occluded facial data and should facilitate open research. For achieving the objectives, we propose two datasets: 1) \textit{\Real}~contains real-world challenging facial occlusions such as beards, food items, hands, sunglasses, and 2) \textit{\Synthetic}~consists of tough natural occlusions such as mustaches, spectacles. Furthermore, to validate the efficacy of our model against noisy cases, we construct a 3) \textit{Noisy} variant of CelebA-test dataset~\cite{liu2015deep}. 
Please refer to the Supplementary for more details about the training and testing datasets.
\\
\noindent\textbf{\Real~Dataset}: To facilitate occlusion robust $3$D face reconstruction model evaluation on challenging real-world data, we introduce a new testing set \textbf{Rea}l-World \textbf{Ch}allenging \textbf{Occ}lusion (\textbf{\Real})~consisting of $550$ face images gathered from various open sources. In our dataset, we have $11$ images of each subject in the set of $50$ subjects such that $10$ images of a subject are occluded, and $1$ image is clean. The occluded and clean facial images are unpaired (captured under different image acquisition environments). These images cover a range of tough facial occlusions, e.g., beards, hands, masks, sunglasses, mustaches, and foods (e.g., Fig.~\ref{fig:data} (a)).
Moreover, we provide $5$ facial landmark coordinates to facilitate cropping and alignment, if needed. However, due to occlusions, $331$ occluded face images failed to be
detected by dlib~\cite{king2009dlib} to produce landmark coordinates. Therefore, we manually labeled the landmark coordinates of these facial images.\\
\noindent\textbf{\Synthetic~Dataset}: We also introduce a novel synthetic occlusion-based test set, \textbf{Syn}thetic \textbf{Ch}allenging \textbf{Occ}lusion (\textbf{\Synthetic}) dataset, to evaluate the performance of occlusion robust $3$D face networks. The dataset contains $550$ face images of $50$ subjects such that each subject has $10$ occluded facial images and $1$ non-occluded face. The occluded are generated by overlaying natural occlusions (e.g., turbans, face masks, eye masks, hats, and bangs) on the clean facial images (e.g., Fig.~\ref{fig:data} (b)); thus, we have paired data in the proposed dataset. Also, we provide $5$ facial landmark coordinates to facilitate cropping and alignment of the face images. These facial landmark coordinates are derived using dlib~\cite{king2009dlib}.\\

\section{Experiments}
In this work, unlike several recent approaches~\cite{feng2018joint,zielonka2022towards}, we aim to recover $3$D face shape and texture simultaneously from occluded and noisy monocular face images without posing additional requirements. To achieve this goal, we propose two datasets designed for this problem (Sec.~\ref{sec:dataset}) since there is no publicly available one. For more dataset and implementation details, please refer to the Supplementary.

\subsection{Evaluation Metrics} 
To evaluate the model performance, we use a standard evaluation metric: \textit{perceptual error metric}, which aims at deriving the mean Euclidean L2 Distance between the feature vectors obtained from various face recognition models. The primary focus of the error metric is on obtaining the visual discrepancy between rendered $3$D face and the corresponding $2$D face image. We exploit a total of $2$ high-performing face recognition models in the main paper: FaceNet-$512$~\cite{schroff2015facenet}, and ArcFace~\cite{deng2019arcface}. We detail an algorithm to outline the perceptual error metric-based evaluation procedure on the proposed \Real~and \Synthetic~datasets in Algo.~\ref{alg:one}. Moreover, we evaluate our approach on the standard NoW~\cite{RingNet:CVPR:2019} validation dataset to validate its effectiveness. We also present the perceptual error results from $5$ other popular backbones, the details on the performance of our model on the MICC dataset in the Supplementary.
\begin{algorithm}\footnotesize{
\caption{Evaluation on \Real~and \Synthetic~Datasets}\label{alg:one}
\begin{algorithmic}
\Require Real-World Occluded Face Dataset: $\boldsymbol{\Psi_{R}}\in \mathbb{R}^{50\times 10}$, Synthetically Occluded Face Dataset: $\boldsymbol{\Psi_{S}}\in \mathbb{R}^{50\times 10}$, Clean Face Dataset: $\boldsymbol{\Upsilon} \in \mathbb{R}^{10}$, Projection Function: $\boldsymbol{\zeta}$, Perceptual Network: $\boldsymbol{\upsilon}$, ROGUE: $\boldsymbol{\lambda}$

\Ensure Perceptual Dissimilarities: $\mathcal{L_R}$, $\mathcal{L_S}$
\While{ $i\leq 50$}
\State $\boldsymbol{I_G} \gets \boldsymbol{\Upsilon}[i]$;
\While{ $j\leq 10$}

\State $\boldsymbol{I_{O^{R}}} \gets \boldsymbol{\Psi_{R}}[i][j]$;
\State $\boldsymbol{I_{O^{S}}} \gets \boldsymbol{\Psi_{S}}[i][j]$;
\State $\boldsymbol{C_{G_S}},\boldsymbol{C_{G_T}},\boldsymbol{C_{G_E}},\boldsymbol{C_{G_I}},\boldsymbol{C_{G_P}} \gets \boldsymbol{\lambda}(\boldsymbol{I_G})$;
\State $\boldsymbol{C_{O_S^{R}}},\boldsymbol{C_{O_T^{R}}},\boldsymbol{C_{O_E^{R}}},\boldsymbol{C_{O_I^{R}}},\boldsymbol{C_{O_P^{R}}} \gets \boldsymbol{\lambda}(\boldsymbol{I_O^{R}})$;

\State $\boldsymbol{C_{O_S^{S}}},\boldsymbol{C_{O_T^{S}}},\boldsymbol{C_{O_E^{S}}},\boldsymbol{C_{O_I^{S}}},\boldsymbol{C_{O_P^{S}}} \gets \boldsymbol{\lambda}(\boldsymbol{I_O^{S}})$;
\State Update $\boldsymbol{C_{O_E^{R}}}\gets \boldsymbol{C_{G_E}}$, $\boldsymbol{C_{O_I^{R}}}\gets \boldsymbol{C_{G_I}}$, $\boldsymbol{C_{O_P^{R}}}\gets \boldsymbol{C_{G_P}}$;
\State $\boldsymbol{I_{O^{R'}}}\gets \boldsymbol{\zeta}(\boldsymbol{C_{O_S^{R}}},\boldsymbol{C_{O_T^{R}}},\boldsymbol{C_{O_E^{R}}},\boldsymbol{C_{O_I^{R}}},\boldsymbol{C_{O_P^{R}}}, \boldsymbol{I_G})$;
\State $\boldsymbol{I_{O^{S'}}}\gets \boldsymbol{\zeta}(\boldsymbol{C_{O_S^{S}}},\boldsymbol{C_{O_T^{S}}},\boldsymbol{C_{O_E^{S}}},\boldsymbol{C_{O_I^{S}}},\boldsymbol{C_{O_P^{S}}}, \boldsymbol{I_G})$;
\State $\mathcal{L}_{R}\gets \boldsymbol{\upsilon}(\boldsymbol{I_{G}}, \boldsymbol{I_{O^{R'}}})$;
\State $\mathcal{L}_{S}\gets \boldsymbol{\upsilon}(\boldsymbol{I_{G}}, \boldsymbol{I_{O^{S'}}})$;
\State $j\gets j+1$
\EndWhile
\State $i\gets i+1$
\EndWhile

\end{algorithmic}
}
\end{algorithm}

\subsection{Experimental Results}
\begin{figure*}[!ht]
    \centering
    \includegraphics[width=0.9\textwidth]{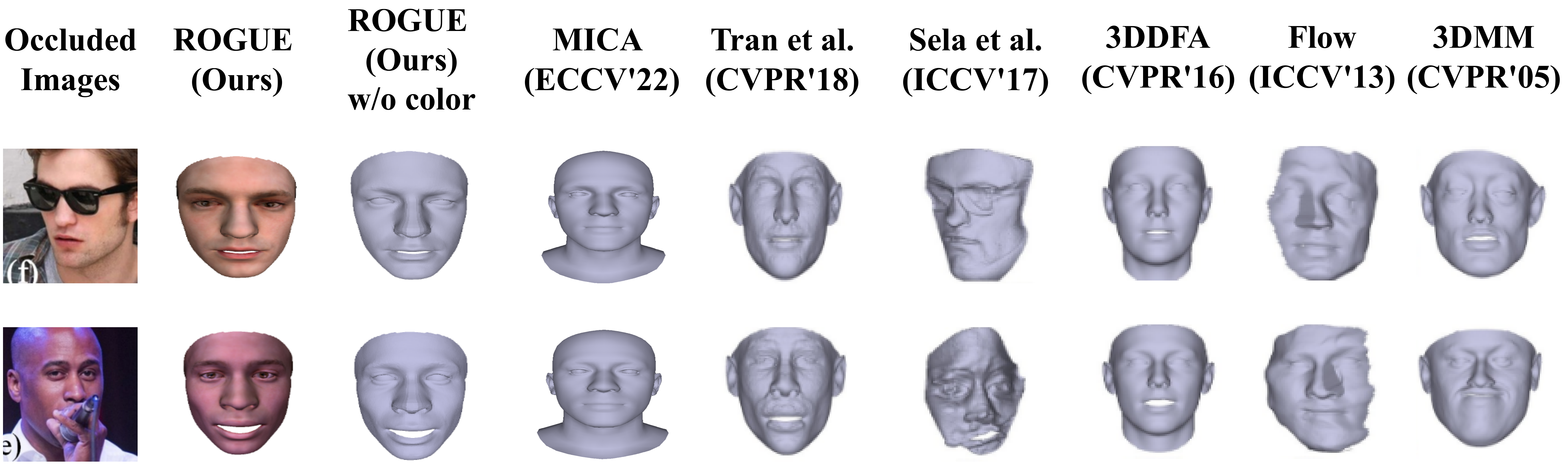}
       \caption{A qualitative comparison of our method with various methods for the case of real-world occlusions. Our results show improved reconstructed $3$D faces. }
    \label{fig:others}
\end{figure*}

\begin{figure}[!ht]
    \centering
    \includegraphics[width=\textwidth]{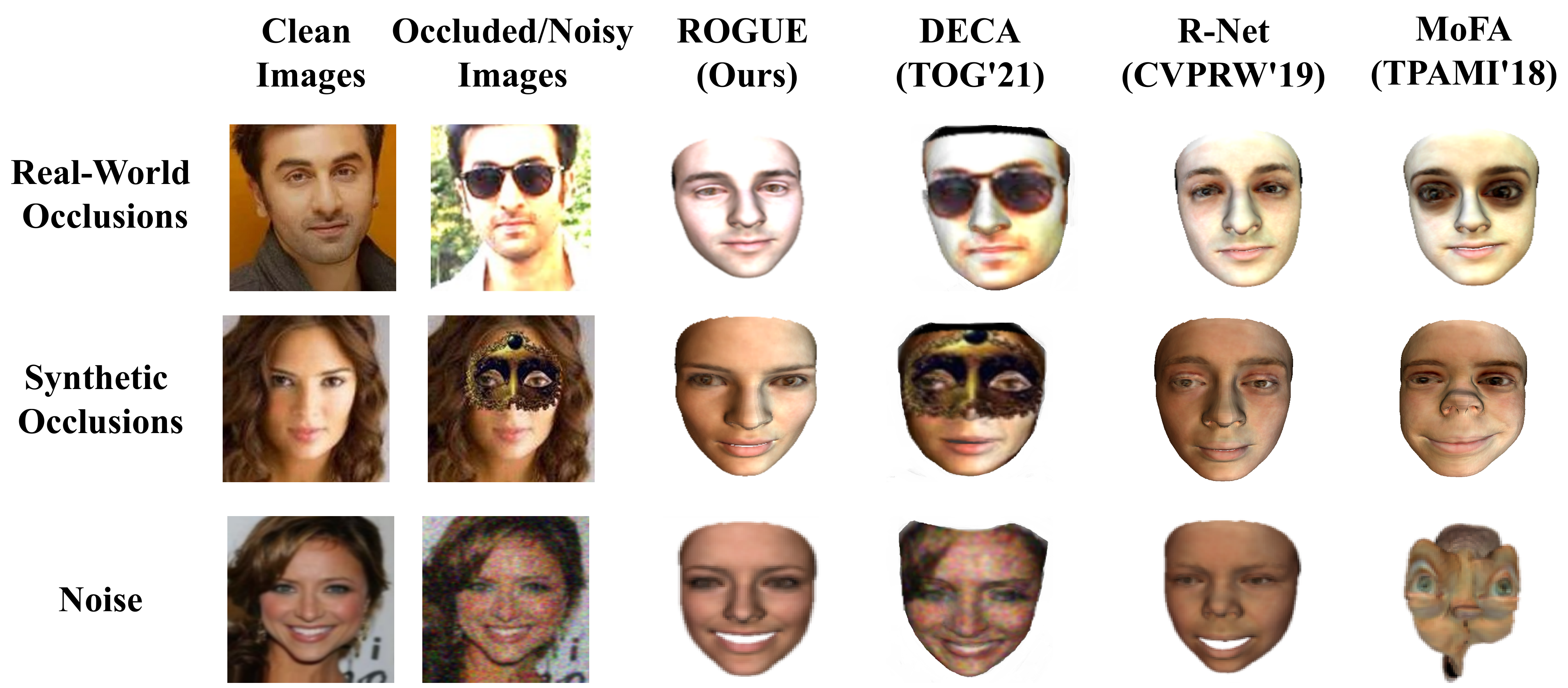}
    \caption{A qualitative comparison of our method with DECA, R-Net and MoFA on the \textit{\Real}, \textit{\Synthetic} and \textit{noisy} datasets. Our results show a significant improvement in the reconstructed $3$D faces. Note that DECA's meshes are cropped for clear comparison.}
    \label{fig:real-MoFAandR-Net}
\end{figure}

\noindent\textbf{Qualitative Evaluation:} We show the qualitative efficacy of our method on: 1) the \textit{\Real} set, 2) the \textit{\Synthetic} set, and 3) the \textit{noisy face} set.
For this purpose, we compare our results with the several latest state-of-the-art methods. $3$DMM~\cite{romdhani2005estimating}, Flow~\cite{hartley2003multiple}, $3$DDFA~\cite{zhu2016face}, Sela et al.~\cite{sela2017unrestricted}, Tran et al.~\cite{tran2018extreme}, and MICA~\cite{zielonka2022towards} are the occlusion robust $3$D face shape reconstruction methods. MoFA~\cite{tewari2018high}, R-Net~\cite{deng2019accurate}, and DECA~\cite{feng2021learning} proposed to reconstruct $3$D face shape and texture simultaneously from occluded monocular face images. Motivated by this, we break our comparisons into two categories: 1) comparison with shape recovery-focused methods (Fig.~\ref{fig:others}), and 2) comparison with texture (along with shape) recovery-based methods (Fig.~\ref{fig:real-MoFAandR-Net}). In Fig.~\ref{fig:others}, our method demonstrates better shape recovery from occluded images than \textit{most} SOTA methods. These approaches focus on recovering shape, whereas texture estimation is beyond the scope of these methods. Moreover, Fig.~\ref{fig:real-MoFAandR-Net} shows that our reconstructed $3$D faces are visually closer to clean images compared to DECA, R-Net, and MoFA. Note that DECA aims at wrapping the input images to the recovered $3$D face shapes by estimating the UV texture maps, thus reproducing occlusions on the $3$D faces. Besides, unlike SOTA approaches, our method simultaneously focuses on recovering occlusion robust shape and texture to improve the visual similarity with the non-occluded facial images.
\noindent\textbf{Quantitative Analysis:} The state-of-the-art methods~\cite{romdhani2005estimating,hartley2003multiple,zhu2016face,sela2017unrestricted,tran2018extreme,zielonka2022towards} focus on reconstructing occlusion-aware $3$D face shapes, whereas the issue of robust texture recovery is not addressed by these methods. Therefore, these methods fail to perform well on the perceptual error metric. As a result, we compare our method with MoFA, R-Net, and DECA, which reconstruct both $3$D face shape and texture. Our quantitative results (Table~\ref{tab:ReaChOcc}) show better perceptual similarity for the reconstructed $3$D face than these approaches. The proposed method reduces the perceptual error by a large margin of $\mathbf{23.8\%}$ (from $1.237$ to $0.943$) compared to MoFA on \Real. In addition, our approach reduces $\mathbf{9.8\%}$ (from $1.045$ to $0.943$) and $\mathbf{14.1\%}$ (from $1.097$ to $0.943$) the perceptual errors for R-Net and DECA, respectively. On the \Synthetic~dataset, our proposed method shows a large reduction of $\mathbf{26.4\%}$ (from $1.195$ to $0.879$) compared to MoFA. In addition, the proposed approach reduces $\mathbf{8.0\%}$ (from $0.955$ to $0.879$) and $\mathbf{7.6\%}$ (from $0.951$ to $0.879$) the perceptual errors with regard to R-Net and DECA, respectively. Finally, for the noisy variant, our method reduces the perceptual errors by a large margin of $\mathbf{22.7\%}$ (from $1.245$ to $0.963$) compared to MoFA. Moreover, our approach reduce $\mathbf{17.1\%}$ (from $1.161$ to $0.963$) and $\mathbf{17.5\%}$ (from $1.167$ to $0.963$) of the perceptual errors compared to R-Net and DECA, respectively. All these results demonstrate the efficacy of the proposed approach. It is worth noting that DECA estimates occlusion robust $3$D face shape, whereas robust texture estimation is beyond its scope; thus, perceptual error evaluation for DECA (Table~\ref{tab:ReaChOcc}) is performed only to emphasize the\textit{ necessity of occlusion robust $3$D texture reconstruction}. We also evaluate our model on the standard NoW~\cite{RingNet:CVPR:2019} validation set. NoW derives the scan-to-mesh distance between the ground truth scan and the predicted meshes. 
It is worth noting that our approach focuses on producing robust texture and shape simultaneously, but the performance on the shape-specific (i.e., not evaluate texture accuracy) NoW dataset (Table~\ref{table:Now}) is still comparable to SOTA methods like DECA.
\begin{table*}[ht]
  
    \centering
  \resizebox{\columnwidth}{!} { \begin{tabular}{ccccc c  cccc}
  \toprule
    & & \multicolumn{2}{c}{\Real}&& \multicolumn{2}{c}{\Synthetic}&& \multicolumn{2}{c}{Noise}\\
    \cmidrule{3-4} \cmidrule{6-7} \cmidrule{9-10}
         Methods&  &FaceNet-$512$&ArcFace&&FaceNet-$512$&ArcFace&&FaceNet-$512$&ArcFace\\
         \midrule
         MoFA (\textbf{TPAMI' 18})&  & $1.237\pm 0.141$& $1.313\pm 0.114$&& $1.195\pm 0.126$&$1.284\pm 0.150$&&$1.245\pm 0.171$&$1.250\pm 0.274$\\
        R-Net (\textbf{CVPRW' 19}) &  &$1.045\pm 0.173$&$1.188\pm 0.171$&&$0.955\pm 0.187$&$1.131\pm 0.194$&&$1.161\pm 0.253$&$1.221\pm 0.217$\\
         DECA (\textbf{TOG' 21}) &  &$1.097\pm 0.176$&$1.196\pm 0.176$&&$0.951\pm 0.184$&$1.061\pm 0.210$&&$1.167\pm 0.295$&$1.170\pm 0.298$\\
        \midrule
        \MainShort~(\textbf{Ours}) & &$\mathbf{0.943\pm 0.187}$ &$\mathbf{1.025\pm 0.168}$&&$\mathbf{0.879 \pm 0.174}$&$\mathbf{0.983\pm 0.186}
        $&&$\mathbf{0.963\pm 0.185}$&$\mathbf{1.017\pm 0.146}$\\ 

        \bottomrule
                           
    \end{tabular}}
 \caption{A quantitative comparison of the perceptual distance using the mean euclidean L2 distance metric with other approaches on the proposed \Real, \Synthetic~and noisy~datasets, where the error numbers are the lower, the better. }  \label{tab:ReaChOcc}
\end{table*} 
\begin{table}[ht]
{
\resizebox{\columnwidth}{!}{
\scriptsize{\begin{tabular}{*{1}{m{0.18\textwidth}}*{1}{m{0.04\textwidth}}*{1}{m{0.04\textwidth}}*{1}{m{0.04\textwidth}}*{1}{m{0.45\textwidth}}}\midrule
\multicolumn{2}{c}{\textbf{\noindent NoW Evaluation (Non-Metrical)}}&&
&\multirow{1}{*}{\includegraphics[width=0.5\textwidth,height=0.16\textheight]{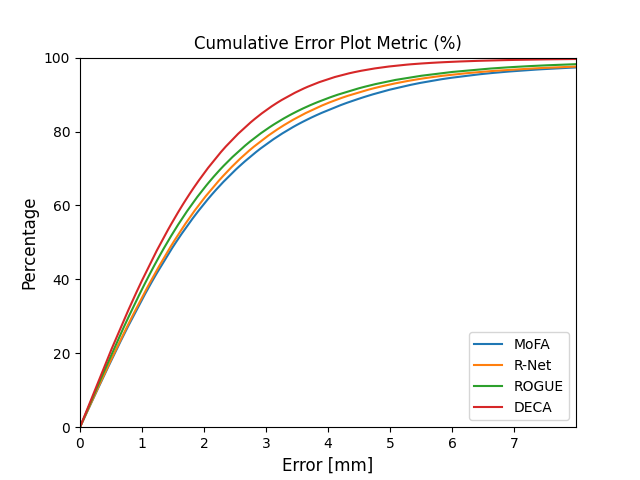}}   \\
\\
\\

\cmidrule{1-4}
\centering{Methods} &\centering{median} &\centering{mean} &\centering{std}&\\
\cmidrule{1-4}
	\centering{MoFA (\textbf{TPAMI' 18})}   &  \centering{$1.547$}     & \centering{$2.228$} & \centering{$2.567$}&\\
	\centering{R-Net (\textbf{CVPRW' 19})}  &     \centering{$1.505$}  &\centering{$2.133$} & \centering{$2.485$} &\\
\centering{\MainShort~(\textbf{Ours})}     & $1.408$     & $1.978$ & $2.221$& \\
\centering{DECA (\textbf{TOG' 21}) }& $1.308$     & $ 1.635$ & $1.407$& \\
\cmidrule{1-4}
\\
\\

\bottomrule
\end{tabular}}}}
\caption{(Left) A quantitative evaluation on the NoW validation dataset. (Right) In the plot, the x-axis shows the scan-to-mesh distance error (in mm), whereas the y-axis displays the cumulative percentage such that the higher the curve, the better the shape-based accuracy. It is worth noting that the expressions are not set to be neutral during evaluation.}
\label{table:Now}
\end{table}


\noindent\textbf{More Discussions}: 
Due to the page limit, please refer to the Supplementary for 1) the details on the testing and training datasets, 2) implementation details, 3) more comparisons with other methods, 4) ablation studies for the cases of occlusions and noisy images, and 5) the discussions of potential negative societal impact and limitations.

\section{Conclusions and Future Work}
In this work, we presented two occluded face datasets, \textbf{\Real}~and \textbf{\Synthetic}, containing various challenging real-world and synthetic occlusion-based face images for robustness tests. Moreover, we proposed a novel \textit{\MainFrameworkBold~(\textbf{\MainShort})} framework to address the problem of occlusions and noise in the face image for monocular $3$D face reconstruction in a self-supervised manner. More specifically, we trained the \textbf{\GuidePipeline} to guide the \textbf{\RobustPipeline} to see through occlusions (e.g., irrespective of the occlusion colors, shapes, and spatial locations) and noise in the face image. Our experiments showed that our model outperforms the current state-of-the-art methods by large margins (e.g., a reduction of $\mathbf{23.8\%}$ for real-world occlusions, $\mathbf{26.4\%}$ for synthetic occlusions, and $\mathbf{22.7\%}$ for the noise in the face images). 
For future work, we aim at even fewer training dependencies. 
For example, we plan to waive the requirement of the \GuidePipeline~by empowering the \RobustPipeline~to self-estimate 
the probable non-occluded $3$D faces that enable the model to gain robustness against the occlusions.
\section*{Acknowledgements}
This work was supported in part by the Higher Education Sprout Project of the National Yang Ming Chiao Tung University, Ministry of Education, and Ministry of Science and Technology (MOST-110-2634-F-A49 -006), Taiwan.

{\small
\bibliography{egbib.bib}
}
\newpage\section*{Supplementary}
In the supplementary, we would like to provide more technical details, experiments, and discussions about the limitations and societal impact.
\section{More Technical Details}
In this section, we detail the preliminaries of the proposed method (refer to Section~\ref{prelim}). Further, in Section~\ref{sec:Main_}, we present the robustifying guidance using \GuidePipeline, which aids the \RobustPipeline~in addressing the issue of occlusions and noise in the face images in a self-supervised manner.
\subsection{Preliminaries: Monocular 3D Face Reconstruction}\label{prelim}
\footnotetext{\hspace{-2.1em}$^\ast$Work partially done during the research internship at MediaTek Inc., Taiwan.}
\footnotetext{\hspace{-2.1em}$^\ast$$^\ast$Work partially done during MediaTek Inc., Taiwan.}
In this section, the preliminaries of the proposed approach are introduced, such as $3$DMM~\cite{blanz1999morphable}, illumination assumptions, and $3$D face projection, which are crucial to address the problem of $3$D reconstruction from monocular face images.\\
\textbf{{3}D Morphable Model ({3}DMM)}: In $3$DMM, a set of geometry and texture coefficients lead to the formation of a $3$D face. Thus the formulas of geometry vector $\mathbf{M}$ and texture vector $\mathbf{T}$ for the $3$DMM model are stated as follows:

\begin{align}\label{eq:3DMM}
 	\mathbf{M} = \overline{\mathbf{M}} + \mathbf{s}\mathbf{B_s}+ \mathbf{e}\mathbf{B_e}, \hspace{2em}
 	\mathbf{T} = \overline{\mathbf{T}} + \mathbf{t}\mathbf{B_t}.
 \end{align}
 
A linear combination of $\mathbf{B_s}\in\mathbb{R}^{3N\times 80}$ and $\mathbf{B_e}\in\mathbb{R}^{3N\times 64}$ (subsets of Principal Component Analysis basis for shape and expression) with the predicted shape parameter $\mathbf{s}=[s_1,\cdots s_{80}]$ and expression parameter $\mathbf{e}=[e_1,\cdots e_{64}]$ respectively morphs the mean $3$D face geometry $\overline{\mathbf{M}}\in\mathbb{R}^{3N}$ (refer to Eq.~\ref{eq:3DMM}). Similarly, texture morphing is facilitated by adding the mean texture $\overline{\mathbf{T}}\in\mathbb{R}^{3N}$ to the linear combination of texture basis vector $\mathbf{B_t}\in\mathbb{R}^{3N\times 80}$ and predicted texture parameter  $\mathbf{t}=[t_1,\cdots t_{80}]$. The vectors $\mathbf{B_t}$, $\overline{\mathbf{T}}$, $\mathbf{B_s}$, and $\overline{\mathbf{M}}$ are obtained from the Basel Face Model~\cite{paysan20093d} whereas we acquire $\mathbf{B_e}$ from the Facewarehouse model~\cite{cao2013facewarehouse}, following~\cite{guo2018cnn}. Note that we, as in~\cite{deng2019accurate}, preclude the ear and neck regions; thus our mesh contains $N = 36$K vertices.

The $3$D face illumination is represented using Spherical Harmonics by assuming a \textit{Lambertian} surface reflectance~\cite{deng2019accurate}. 
To obtain the face image, the $3$D face coordinates are mapped to the screen by assuming a pinhole camera under full perspective projection, as in~\cite{deng2019accurate}.
We represent the $3$D face illumination using Spherical Harmonics by assuming a \textit{Lambertian} surface reflectance. 

\subsection{\GuidePipeline}\label{sec:Main_}
One of the main goals is learning reliable $3$DMM coefficients with \textit{the least} supervision and dependencies. Inspired by R-Net~\cite{deng2019accurate} which contains comparatively fewer dependencies, we propose the \textbf{Self-Supervised \GuidePipeline} to learn the coefficients $\boldsymbol{C_G}$ by exploiting the cycle-consistency in a self-supervised manner (see main paper for details). All the components of the \textit{\GuidePipeline} are presented as follows:

\noindent\textbf{Obtaining 3D Face Alignment}: 
The first consistency we aim to maintain is the face alignment between the guiding image $I_G$ and the reconstructed image $I_{G'}$, which is achieved by reducing the discrepancy between landmark coordinates of the faces. We represent the discrepancy using the \textit{Landmark Loss} $\mathcal{L}_K$ as follows:
Following R-Net, we obtain the alignment between the $2$D non-occluded noise-free face image and the corresponding $3$D face projection by reducing the discrepancies between the $68$ landmark coordinates of the faces.
\begin{align}\label{eq:lndmrk}
     \mathcal{L}_K = || L_G-{L_{G'}} ||.
 \end{align}
where $L_G$ and ${L_{G'}}$ denote a set of $68$ landmark coordinates of $2$D face image and the rendered counterpart, respectively, and $||.||$ represents the L2 norm.

\noindent\textbf{Obtaining Photometric Consistency}: 
To reach the goal of fewer dependencies, we directly regress the pixels of the rendered $3$D face ($I_{G'}$) over the corresponding guiding face image ($I_G$) and obtain the pixel-wise consistency between them by the \textit{Guiding Photometric Loss} $\mathcal{L}_{GP}$ as follows:
\begin{align}\label{eq:GP}
    \mathcal{L}_{GP} = || I_G-I_{G'} ||.
\end{align}
It is worth noting that unlike R-Net~\cite{deng2019accurate}, we relax the requirement of skin masks as additional dependencies.

\noindent\textbf{Obtaining Perceptual Loss}: 
In addition to image-level information, reducing the feature-level discrepancy is critical to obtaining the perceptual accuracy of $3$D faces. Thus, we adopt the \textit{Perceptual Loss} $\mathcal{L}_{P}$ as follows:
\begin{align}\label{eq:PA}
    \mathcal{L}_{P} = 1-\frac{\langle\theta, \theta'\rangle}{||\theta|| ||\theta'||}.
\end{align}
where $\theta$ and $\theta'$ are the feature representations obtained from the pre-trained FaceNet model~\cite{schroff2015facenet} for target image $I_G$ and the corresponding rendered face $I_{G'}$, respectively.

\noindent\textbf{Regularization}: To ensure the plausible face geometry and texture of the reconstructed $3$D face, we adopt the \textit{Regularization} term $\mathcal{L}_{R}$, which enforces the coefficients to follow the (normal) distribution of $3$DMM, as follows:
\begin{align}\label{eq:reg}
    \mathcal{L}_{R} = w_{s}||\mathbf{s}||+w_{t}||\mathbf{t}||+w_{e}||\mathbf{e}||,
\end{align}
where $w_{s}, w_{t}$ and $w_{e}$ are the weights associated with shape $\mathbf{s}$, texture $\mathbf{t}$, and expression $\mathbf{s}$ coefficients, respectively.

\noindent\textbf{Overall Loss for \GuidePipeline}: 
The overall loss function $\mathcal{L}_{guide}$ for the \textit{Self-Supervised \GuidePipeline}~can be expressed below:
\begin{align}\label{EQ:guide}
    \mathcal{L}_{guide} = \alpha_K \mathcal{L}_{K}+\alpha_{GP} \mathcal{L}_{GP}+\alpha_{P} \mathcal{L}_{P}+\alpha_{R} \mathcal{L}_{R}.
\end{align}
where $\alpha_K, \alpha_{GP}, \alpha_{P}$, and $\alpha_{R}$ are the weights for landmark loss (Eq.~\eqref{eq:lndmrk}), guiding photometric loss (Eq.~\eqref{eq:GP}), perceptual loss (Eq.~\eqref{eq:PA}) and regularization term (Eq.~\eqref{eq:reg}), respectively.
For simplicity, the notation of the image index is ignored throughout the whole paper. 

\section{More Experimental Details}
In this section, we detail the datasets and the corresponding variations deployed for training the proposed \textit{Self-Supervised} \textit{\textbf{RO}bustifying} \textit{\textbf{GU}idanc\textbf{E}} \textit{(\textbf{\MainShort})} framework (refer to Sec.~\ref{sec:data}). In addition, we present the experimental details, such as network architecture, weights for the losses, etc., in Sec.~\ref{sec:network}. Finally, more ablation studies and experiments are shown in Sec.~\ref{sec:ablation} and \ref{sec:more_results}, respectively.

\subsection{Datasets}\label{sec:data}
\begin{figure}
     \centering
    \includegraphics[width=0.9\textwidth]{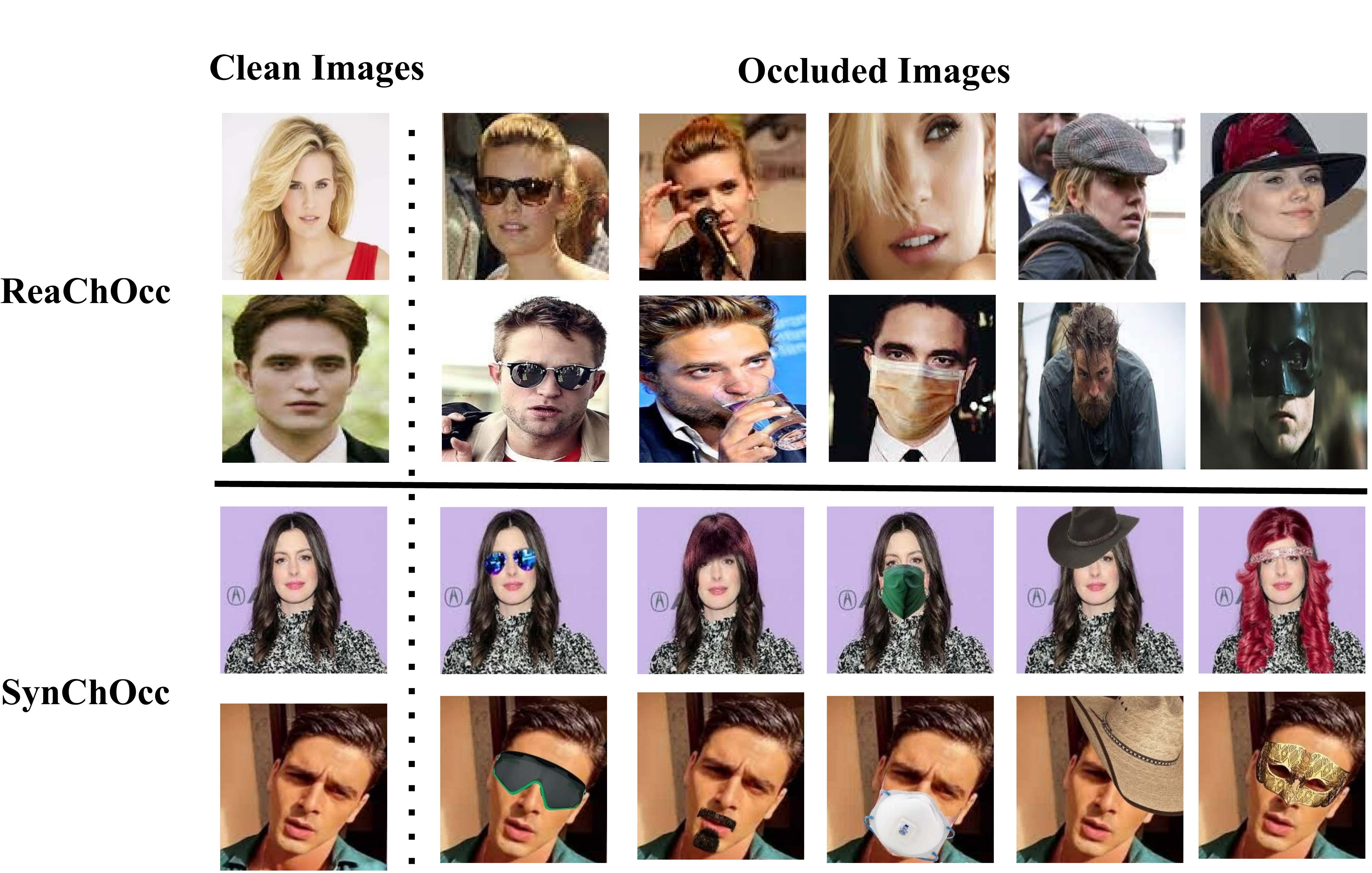}
    \caption{A demonstration of the samples from the proposed \Real~and \Synthetic~datasets. Our \Real~contains unpaired clean images, whereas \Synthetic~provides paired clean images for comparison.}
    \label{fig:data_}
\end{figure}

Procuring $3$D ground-truth data is difficult due to privacy concerns and monetary issues. 
To validate the robustness against occlusions and noise, we exploit the proposed 1) \Real~dataset, 2) \Synthetic~dataset, and build a variant of the test set of CelebA~\cite{liu2015deep} 3) \textit{noisy face} set. A sample of the proposed \Real~and \Synthetic~datasets is shown in Fig.~\ref{fig:data_}.
For quantitative evaluation, we use a total of $7$ high-performing face recognition models (the results of $2$ are shown in the main paper and the remaining in supplementary) to determine the perceptual accuracy between the input face image and the corresponding rendered face. Our occlusion-based testing (real-life) and training (occluded with random shapes) images belong to different domains. Therefore, our method is distinct from the conventional data augmentation-based techniques, where the domains of test and training data are needed to be the same. 

\subsection{Implementation Details}\label{sec:network}
The proposed \textit{\MainFrameworkBold~(\textbf{\MainShort})}~framework contains a coefficient network and two discriminators to facilitate the overall learning of the model. The coefficient network exploits ResNet-50 as backbone architecture with a modified classification layer by $257$ nodes. The face images in the dataset are cropped, aligned (using the method in~\cite{chen2016supervised}), and reshaped
to size $224\times 224$. These images serve as the input to our model. We opt for a batch of $5$ for each case: clean images, occluded faces, and noisy face images. Thus, the proposed network is trained with a net batch size of $15$. In addition, we exploit linear discriminators with $3$ fully-connected layers containing $257$, $124$, and $2$ nodes, respectively, in the \RobustPipeline. Our model is initialized with ImageNet weights~\cite{russakovsky2015imagenet}. In addition, an Adam optimizer~\cite{kingma2015adam} is deployed for training the model with an initial learning rate of $10^{-4}$ for the coefficient network and $10^{-8}$ for the discriminators. The proposed model contains the  \textit{\GuidePipeline}~and the \textit{\RobustPipeline}, where the weights associated with the losses in \GuidePipeline~are $\alpha_K = 1.6\times 10^{-3}, \alpha_{GP}=1.92, \alpha_P=0.2$ and $\alpha_R=3\times 10^{-4}$ (as in R-Net), and the weights for \RobustPipeline~are $\beta_O=1.92, \beta_N=1.92$ and $\beta_C= 10^{-3}$ (please refer to Sec.~\ref{sec:ablation} for more ablation experimental results).

\begin{table*}

    \centering
  \resizebox{\columnwidth}{!} { \begin{tabular}{cccccccc c c}
    \toprule
        &  &&&&&\multicolumn{3}{c}{Reported Robustness}
        \\\cmidrule{7-9}
                Methods&  &{Learning}&\thead{Require \\Real Occluded Input}&{External Aid}& &\thead{Real-world\\ Occlusions}&\thead{Synthetic\\ Occlusions}  &Noise\\
         \midrule
           MoFA (TPAMI'18)  &   &N/A &$\checkmark$ &N/A &&$\checkmark$&$\times$&$\times$\\
          $3$DMM (CVPR'05) &   &N/A &$\checkmark$ &N/A &&$\checkmark$&$\times$&$\times$\\
          Flow (Cambridge Press'03) &   & N/A& $\checkmark$& N/A&&$\checkmark$&$\times$&$\times$\\
          $3$DFFA (CVPR'16) &   &N/A &$\checkmark$ &N/A &&$\checkmark$&$\times$&$\times$\\
          Sela et al. (ICCV'17) & &N/A  &$\checkmark$  &N/A &&$\checkmark$&$\times$&$\times$\\
         Tran et al. (CVPR'18) &   &Weakly-Supervised &$\checkmark$ & Bump Map&&$\checkmark$&$\times$&$\times$\\
           
        R-Net (CVPRW'19) &   & Weakly-Supervised & $\checkmark$&Attention Mask &&$\checkmark$&$\times$&$\times$\\
         DECA (TOG'21) &   & Weakly-Supervised& $\checkmark$& Attention Mask&&$\checkmark$&$\times$&$\times$\\
         MICA (ECCV'22) &   & Fully-Supervised& $\checkmark$& Pre-trained ArcFace&&$\checkmark$&$\times$&$\times$\\
        \midrule
        \textbf{\MainShort~(Ours)} &   & Self-Supervised& $\times$& No Aid Required&&$\checkmark$&$\checkmark$&$\checkmark$\\ 

        \bottomrule
                           
    \end{tabular}} 
     	    \caption{A comparison of the proposed method with various state-of-the-art approaches based on the learning scheme, input constraints, the requirement of external aid, and the reported robustness. Unlike SoTA approaches, our method performs well for four challenging data variations, three variants of occlusions, and noise in the images, without posing any dependency. }
    \label{tab:comparison}

\end{table*}

\subsection{More Experimental Results}\label{sec:more_results} 
This section qualitatively compares our method with DECA, MoFA, and R-Net on real-world and synthetic occlusions, and noisy images. Moreover, we provide the quantitative evaluation of our model on $7$ face recognition models: VGG-Face~\cite{cao2018vggface2}, FaceNet~\cite{schroff2015facenet}, FaceNet-$512$~\cite{schroff2015facenet}, OpenFace~\cite{amos2016openface}, DeepFace~\cite{serengil2021tensorflow}, ArcFace~\cite{deng2019arcface} and SFace~\cite{zhong2021sface}. \textit{It is worth noting that we demonstrate the perceptual results from DECA to emphasize the necessity of occlusion robust $3$D texture reconstruction. However, we understand that DECA is not designed for occlusion-robust texture reconstruction.}  We also provide the details on how our work differs from various other approaches in Table~\ref{tab:comparison}.

\subsubsection{Qualitative Evaluation}
\begin{figure}[!ht]
\centering
\includegraphics[width=0.8\columnwidth]{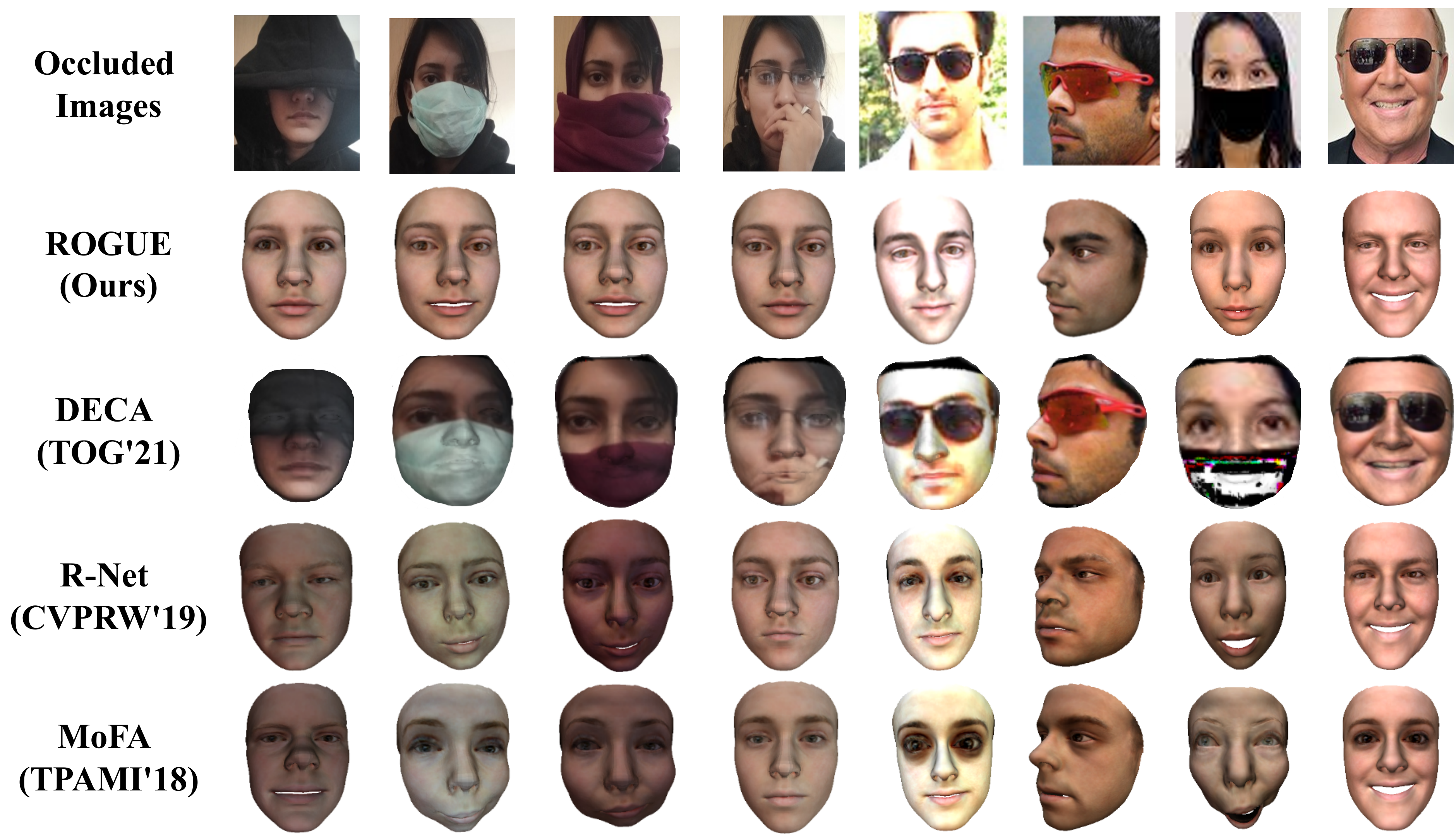}
\caption{A qualitative comparison for different methods on \textit{real-life} occlusions. }
\label{fig:real_}
\end{figure}

\begin{figure}[!ht]
\centering
    \includegraphics[width=0.8\columnwidth]{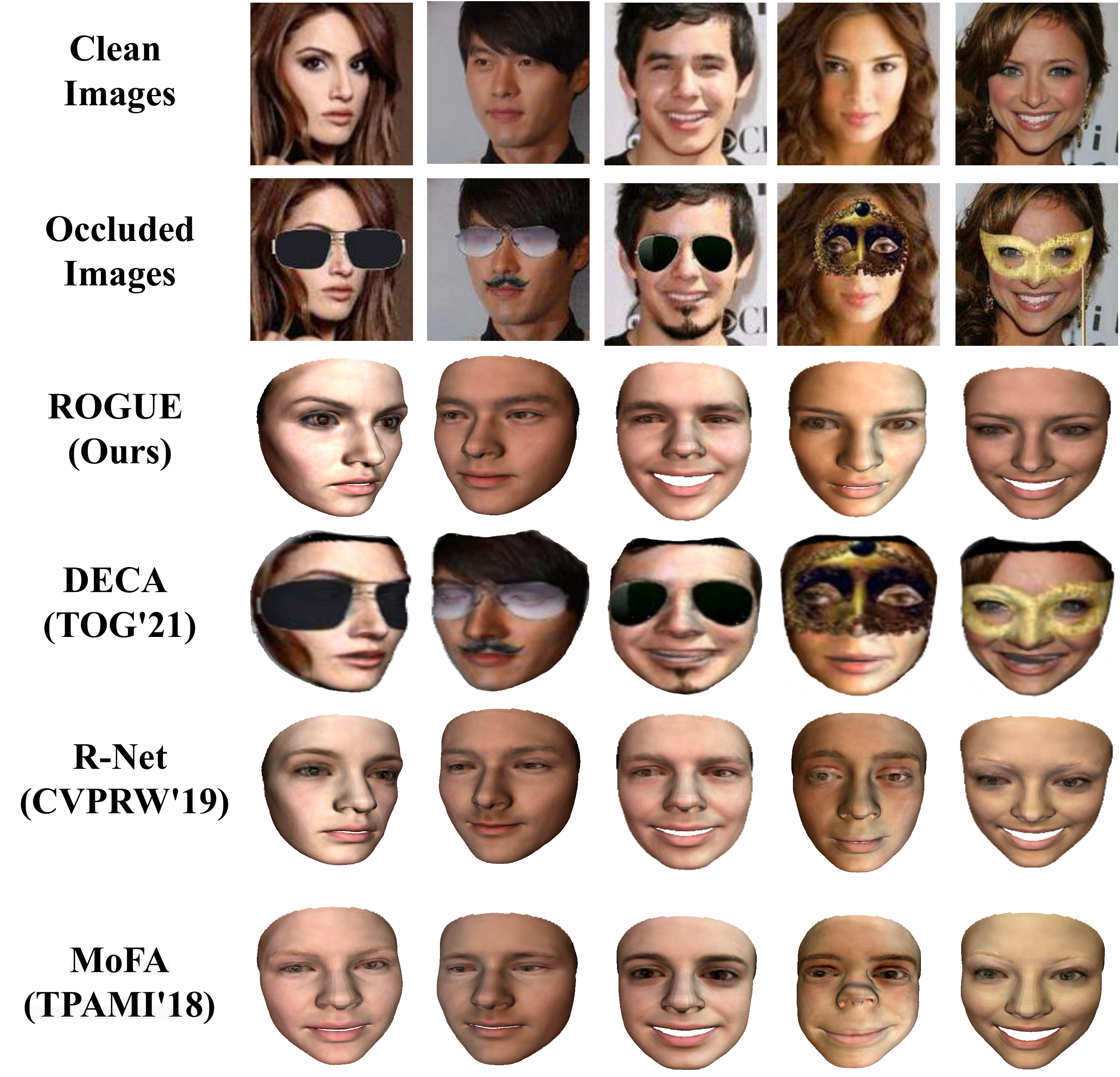}
    \caption{\textit{A qualitative comparison for different methods on \textit{synthetic} occlusions.}}
    \label{fig:Occ_}
\end{figure}

\begin{figure}
\centering
    \includegraphics[width=0.6\columnwidth]{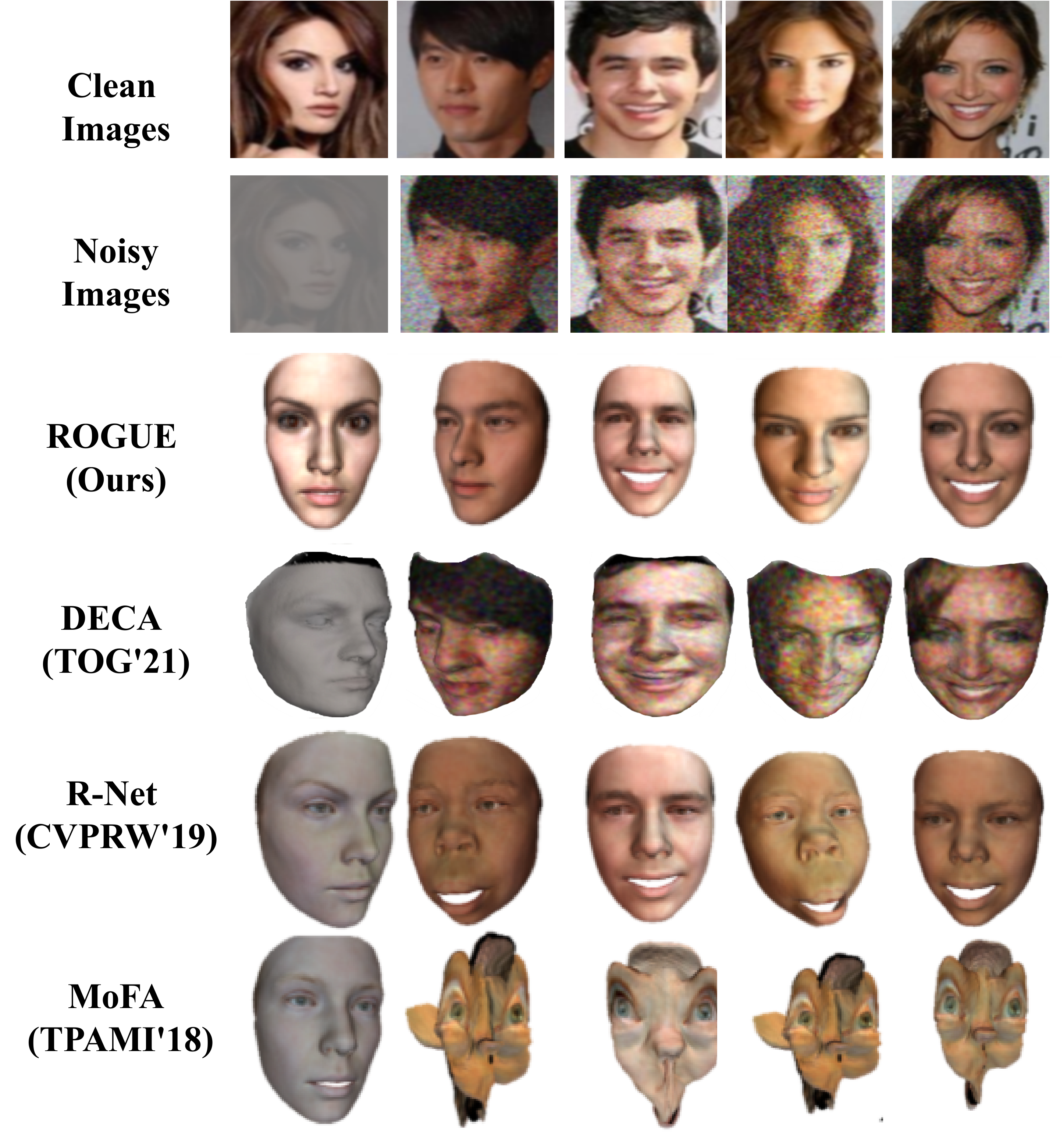}
    \caption{A qualitative comparison for different methods on \textit{noisy images} (various noise types).
    }
    \label{fig:noise}
\end{figure}

To validate the generalization ability of our method, we rigorously test our model on real and synthetic occlusion and noisy scenarios in the main paper. Following the course, in Fig.~\ref{fig:real_}, Fig.~\ref{fig:Occ_}, and Fig.~\ref{fig:noise}, we present various examples to show the robustness of our model toward \textit{unseen} real-life, synthetic occlusions and noisy challenging cases in the face images. It is worth noting that the improvement over MoFA and R-Net is more evident when the occlusion cover more than half of the face and the occlusion is not in skin color. Besides, DECA reproduces occlusion on the $3$D faces, thus following a different line of research. We infer the same for the case of noisy face images.
For a fair comparison, MoFA is re-trained to reconstruct $3$D faces with the same number of face vertices ($N=36K$), and only the trained CNN-based models (e.g., our coefficient network) are used at the inference stage.


\subsubsection{Quantitative Evaluation} 
We investigate the quantitative performance for the \textit{real-world challenging occlusions}, \textit{synthetic challenging occlusions} and \textit{noisy images}. As evident in the qualitative results in the main paper that only three methods, MoFA, R-Net, and DECA, reconstruct both $3$D face shape and texture; thus, we compare the quantitative performance of our method with these three approaches.

\noindent\textbf{1) Real-World Challenging Occlusions:} Our quantitative results in Table~\ref{tab:ReaChOcc_} show better perceptual accuracy for the reconstructed $3$D face than R-Net and MoFA. Our proposed method reduces perceptual errors between the vectors obtained from VGG-Face, FaceNet, FaceNet-$512$, OpenFace, DeepFace, ArcFace, and SFace by a large margin of $\mathbf{17.44\%}$ (from $0.952$ to $0.786$), $\mathbf{21.55\%}$ (from $1.202$ to $0.943$), $\mathbf{25.8\%}$ (from $1.237$ to $0.918$), $\mathbf{11.3\%}$ (from $0.769$ to $0.682$), $\mathbf{8.1\%}$ (from $0.656$ to $0.603$), $\mathbf{21.9\%}$ (from $1.313$ to $1.025$), $\mathbf{13.4\%}$ (from $1.215$ to $1.052$), respectively, compared to MoFA. In addition, the proposed method reduces $\mathbf{5.8\%}$ (from $0.834$ to $0.786$), $\mathbf{8.0\%}$ (from $1.025$ to $0.943$), $\mathbf{12.1\%}$ (from $1.045$ to $0.918$), $\mathbf{11.3\%}$ (from $0.706$ to $0.682$), $\mathbf{3.4\%}$ (from $0.621$ to $0.603$), $\mathbf{13.7\%}$ (from $1.188$ to $1.025$), $\mathbf{6.0\%}$ (from $1.119$ to $1.052$) of the perceptual distance between vectors obtained using VGG-Face, FaceNet, FaceNet-$512$, OpenFace, DeepFace, ArcFace, and SFace, respectively, from R-Net. Moreover, the proposed method reduces $\mathbf{19.4\%}$ (from $0.975$ to $0.786$), $\mathbf{18.3\%}$ (from $1.154$ to $0.943$), $\mathbf{16.3\%}$ (from $1.097$ to $0.918$), $\mathbf{32.2\%}$ (from $1.005$ to $0.682$), $\mathbf{38.3\%}$ (from $0.977$ to $0.603$), $\mathbf{14.3\%}$ (from $1.196$ to $1.025$), $\mathbf{22.4\%}$ (from $1.356$ to $1.052$) of the perceptual distance between vectors obtained using VGG-Face, FaceNet, FaceNet-$512$, OpenFace, DeepFace, ArcFace, and SFace, respectively, from DECA. All the above results show that the proposed approach tackles real-world occlusions more effectively.

\begin{table*}[!ht]

    \centering
  \resizebox{\columnwidth}{!} { \begin{tabular}{ccccccccc}
    \toprule
    & & \multicolumn{7}{c}{Perceptual Error ($\downarrow$)}\\
    \cmidrule{3-9}
         Methods&  &VGG-Face&FaceNet&FaceNet-$512$&OpenFace&DeepFace&ArcFace&SFace\\
         \midrule

          MoFA (TPAMI'18)&  & $0.952\pm 0.129$& $1.202\pm 0.146$& $1.237\pm 0.141$&$0.769\pm 0.189$&$0.656\pm 0.187$&$1.313\pm 0.114$&$1.215\pm 0.119$\\
        R-Net (CVPRW'19) &  &$0.834\pm 0.143$&$1.025\pm 0.195$&$1.045\pm 0.173$&$0.706\pm 0.189$&$0.621\pm 0.180$&$1.188\pm 0.171$&$1.119\pm 0.143$\\
        DECA (TOG'21)&  & $0.975\pm 0.131$& $1.154\pm 0.178$&$1.097\pm 0.176$&$1.005\pm 0.207$&$0.977\pm 0.111$&$1.196\pm 0.176$&$1.356\pm 0.105$\\
        \midrule
        \textbf{\MainShort~(Ours}) & &$\mathbf{0.786 \pm 0.138}$ &$\mathbf{0.943\pm 0.187}$&$\mathbf{0.918\pm 0.171}$&$\mathbf{0.682\pm 0.189}
        $&$\mathbf{0.603\pm 0.176}$&$\mathbf{1.025\pm 0.168}$&$\mathbf{1.052\pm 0.140}$\\ 

        \bottomrule
                           
    \end{tabular}} \caption{A quantitative comparison of the perceptual error metric with other approaches on the proposed \textit{\Real~dataset}, where the error numbers are the lower the better. }
		 
    \label{tab:ReaChOcc_}
 
\end{table*} 

\noindent\textbf{2) Synthetic Challenging Occlusions:}
To further investigate the impact of occlusions, we build the \textit{synthetic occlusion} set \Synthetic~by overlaying natural occlusions at the specific spatial location on clean face images, leading to much more challenging occlusion conditions than the real-world ones.
Synthetic occlusions are hard to be tackled. Thus, we present comprehensive comparisons of the proposed method with the other recent approaches in Table~\ref{tab:syn_}. Our quantitative results show better perceptual accuracy for the reconstructed $3$D face than R-Net and MoFA. Our proposed method reduces perceptual distance between vectors obtained from VGG-Face, FaceNet, FaceNet-$512$, OpenFace, DeepFace, ArcFace, and SFace by a large margin of $\mathbf{15.8\%}$ (from $0.952$ to $0.802$), $\mathbf{23.0\%}$ (from $1.157$ to $0.891$), $\mathbf{26.4\%}$ (from $1.195$ to $0.879$), $\mathbf{2.6\%}$ (from $0.894$ to $0.871$), $\mathbf{0.8\%}$ (from $0.854$ to $0.847$), $\mathbf{23.4\%}$ (from $1.284$ to $0.983$), $\mathbf{18.1\%}$ (from $1.241$ to $1.016$), respectively, compared to MoFA. In addition, the proposed method reduces $\mathbf{2.7\%}$ (from $0.824$ to $0.802$), $\mathbf{21.55\%}$ (from $0.968$ to $0.891$), $\mathbf{8.0\%}$ (from $0.955$ to $0.879$), $\mathbf{11.3\%}$ (from $0.880$ to $0.871$), $\mathbf{1.0\%}$ (from $0.860$ to $0.847$), $\mathbf{13.1\%}$ (from $1.131$ to $0.983$), $\mathbf{12.6\%}$ (from $1.162$ to $1.016$) of the perceptual distance between vectors obtained using VGG-Face, FaceNet, FaceNet-$512$, OpenFace, DeepFace, ArcFace, and SFace, respectively. Furthermore, our approach shows an improvement of $\mathbf{4.2\%}$ (from $0.837$ to $0.802$), $\mathbf{11.1\%}$ (from $1.002$ to $0.891$), $\mathbf{7.6\%}$ (from $0.951$ to $0.879$), $\mathbf{3.9\%}$ (from $0.906$ to $0.871$), $\mathbf{7.6\%}$ (from $0.917$ to $0.847$), $\mathbf{7.4\%}$ (from $1.061$ to $0.983$), $\mathbf{19.5\%}$ (from $1.262$ to $1.016$) in the perceptual distance between vectors obtained using VGG-Face, FaceNet, FaceNet-$512$, OpenFace, DeepFace, ArcFace, and SFace, respectively than DECA.

\begin{table}[!ht]   
    \centering
  \resizebox{\columnwidth}{!} { \begin{tabular}{ccccccccc}
    \hline
    &&\multicolumn{7}{c}{Perceptual Error ($\downarrow$)}\\
    \cmidrule{3-9}
         Methods&  &VGG-Face&FaceNet&FaceNet-$512$&OpenFace&DeepFace&ArcFace&SFace\\
         \midrule

          MoFA (TPAMI'18)&  & $0.952\pm 0.130$& $1.157\pm 0.133$& $1.195\pm 0.126$&$0.894\pm 0.196$&$0.854\pm 0.165$&$1.284\pm 0.150$&$1.241\pm 0.129$\\
        R-Net (CVPRW'19) &  &$0.824\pm 0.155$&$0.968\pm 0.186$&$0.955\pm 0.187$&$0.880\pm 0.195$&$0.860\pm 0.165$&$1.131\pm 0.194$&$1.162\pm 0.170$\\
        DECA (TOG'21)&  & $0.837\pm 0.144$& $1.002\pm 0.197$& $0.951\pm 0.184$&$0.906 \pm 0.201$&$0.917\pm 0.162$&$1.061 \pm 0.210$&$1.262\pm 0.166$\\
        \midrule
        \textbf{\MainShort~(Ours)} &  &$\mathbf{0.802\pm 0.151}$&$\mathbf{0.891\pm 0.179}$&$\mathbf{0.879\pm 0.174}
        $&$\mathbf{0.871\pm 0.195}$&$\mathbf{0.847\pm 0.165}$&$\mathbf{0.983\pm 0.186}$&$\mathbf{1.016\pm 0.168}$\\ 

        \bottomrule
                           
    \end{tabular}}  \caption{A quantitative comparison of the perceptual distance using perceptual error with other approaches on the proposed \textit{\Synthetic~dataset}, where the error numbers are the lower the better. } 
    \label{tab:syn_}

\end{table} 

\noindent \textbf{3) Noisy Faces:} Finally, we investigate the case of noisy face images by introducing various types of noise such as speckle, salt, pepper, Gaussian, etc.
The quantitative results are shown in Table~\ref{tab:noise__}. 
Our proposed method reduces perceptual distance between vectors obtained from VGG-Face, FaceNet, FaceNet-$512$, OpenFace, DeepFace, ArcFace, and SFace by a large margin of $\mathbf{19.6\%}$ (from $0.996$ to $0.801$), $\mathbf{22.9\%}$ (from $1.265$ to $0.976$), $\mathbf{22.7\%}$ (from $1.245$ to $0.963$), $\mathbf{6.8\%}$ (from $0.923$ to $0.860$), $\mathbf{17.2\%}$ (from $0.791$ to $0.655$), $\mathbf{18.6\%}$ (from $1.250$ to $1.017$), and $\mathbf{10.1\%}$ (from $1.260$ to $1.133$) respectively, compared to MoFA.
Moreover, the proposed method reduces $\mathbf{15.3\%}$ (from $0.946$ to $0.801$), $\mathbf{16.0\%}$ (from $1.165$ to $0.979$), $\mathbf{17.1\%}$ (from $1.161$ to $0.963$), $\mathbf{7.3\%}$ (from $0.928$ to $0.860$), $\mathbf{20.7\%}$ (from $0.826$ to $0.655$), $\mathbf{16.7\%}$ (from $1.221$ to $1.017$), $\mathbf{9.7\%}$ (from $1.255$ to $1.133$) of the perceptual distance between vectors obtained using VGG-Face, FaceNet, FaceNet-$512$, OpenFace, DeepFace, ArcFace, and SFace, respectively with regard to R-Net. Furthermore, our approach shows an improvement of $\mathbf{16.2\%}$ (from $0.956$ to $0.801$), $\mathbf{17.6\%}$ (from $1.188$ to $0.979$), $\mathbf{17.5\%}$ (from $1.167$ to $0.963$), $\mathbf{21.0\%}$ (from $1.089$ to $0.860$), $\mathbf{29.0\%}$ (from $0.923$ to $0.655$), $\mathbf{13.1\%}$ (from $1.170$ to $1.017$), $\mathbf{13.6\%}$ (from $1.312$ to $1.133$) of the perceptual distance between vectors obtained using VGG-Face, FaceNet, FaceNet-$512$, OpenFace, DeepFace, ArcFace, and SFace, respectively with regard to DECA.

\begin{table*}[!ht]    
    \centering
  \resizebox{\columnwidth}{!} { \begin{tabular}{ccccccccc}
    \hline
    &&\multicolumn{7}{c}{Perceptual Error ($\downarrow$)}\\
    \cmidrule{3-9}
         Methods&  &VGG-Face&FaceNet&FaceNet-$512$&OpenFace&DeepFace&ArcFace&SFace\\
         \midrule
        
          MoFA (TPAMI'18) &  & $0.996\pm 0.150$& $1.265\pm 0.168$& $1.245\pm 0.171$&$
0.923\pm 0.221$&$0.791\pm 0.175$&$1.250 \pm 0.274$&$1.260\pm 0.112$\\
        R-Net (CVPRW'19)&  &$0.946\pm 0.203$&$1.165\pm 0.250$&$1.161\pm 0.253$&$0.928\pm 0.263$&$0.826\pm 0.207$&$1.221\pm 0.217$&$1.255\pm 0.156$\\
      DECA (TOG'21)&  & $0.956\pm 0.204$&$1.188\pm 0.262$& $1.167\pm 0.295$ &$1.089\pm 0.223$&$0.923\pm 0.151$&$1.170\pm0.298$&$1.312 \pm 0.142$\\
        \midrule
        \textbf{\MainShort~(Ours)} &  &$\mathbf{0.801\pm 0.171}$&$\mathbf{0.979\pm 0.195}$&$\mathbf{0.963\pm 0.185}
        $&$\mathbf{0.860\pm 0.194}$&$\mathbf{0.655\pm 0.181}$&$\mathbf{1.017\pm 0.146}$&$\mathbf{1.133\pm 0.121}$\\ 

        \bottomrule
                           
    \end{tabular}} \caption{
		A quantitative comparison of the perceptual error metric with other approaches on the proposed \textit{noisy variant} of CelebA-test dataset, where the error numbers are the lower the better. }  
    \label{tab:noise__}

\end{table*}

\begin{figure}[!ht]
\center
    \includegraphics[width=0.85\textwidth]{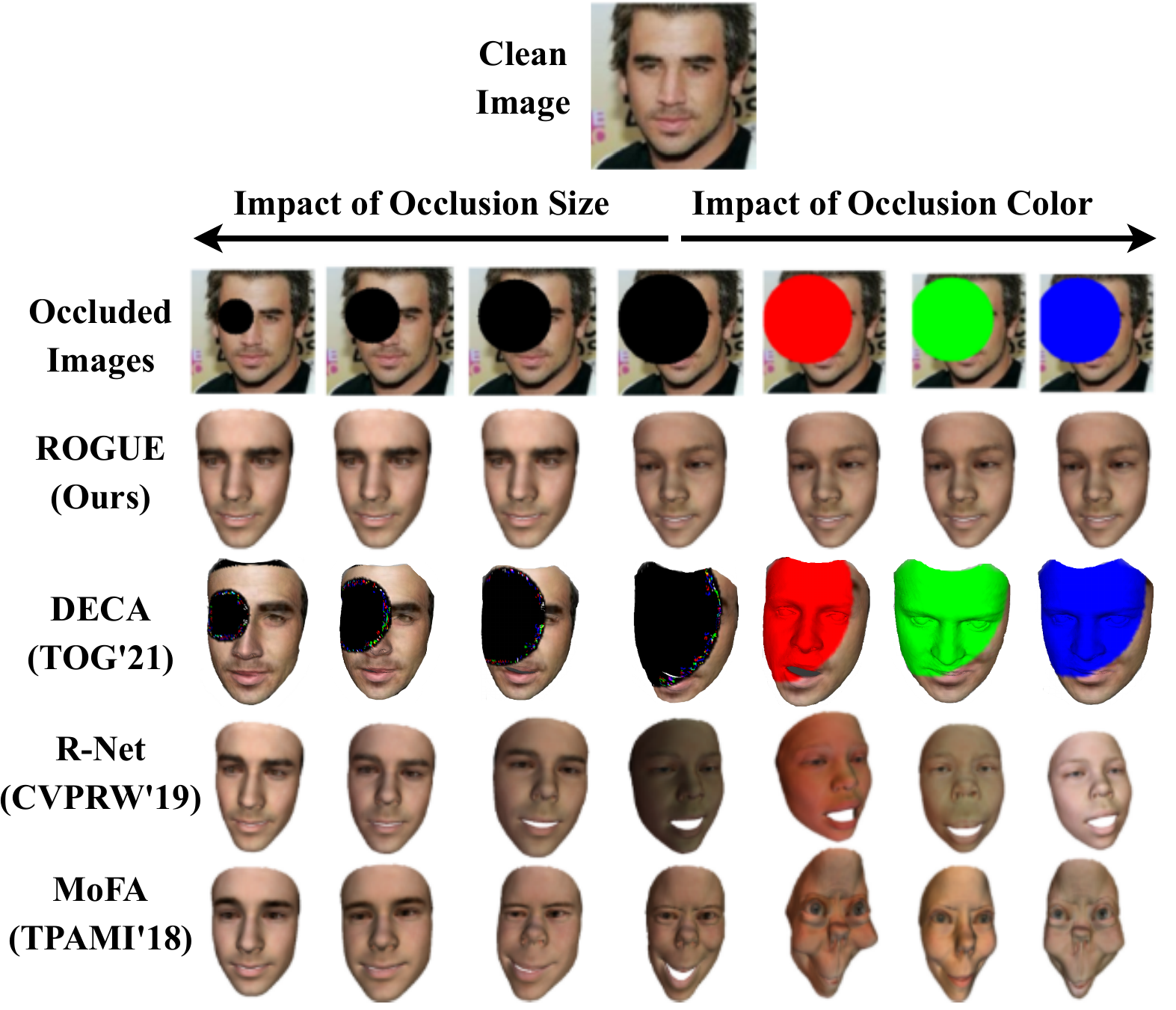}
    \caption{
    A qualitative comparison of the reconstructed 3D faces with other approaches for different \textit{occlusion sizes and colors}. The results show that the proposed method carries better robustness against various colors and sizes of occlusions.
    }
    \label{fig:black_inpainted}
\end{figure}

\subsection{Ablation Studies}\label{sec:ablation}
\noindent\textbf{Impact of Occlusion Color and Size.} 
A critical question also arises: \textit{How do colors and sizes of occlusions affect the reconstructed $3$D faces?} 
To answer this, we present a detailed study with various sizes and colors of facial occlusions across the fixed face image. In Fig.~\ref{fig:black_inpainted}, it is observed that our method is highly robust to the sizes of occlusions. Moreover, the reconstructed $3$D faces are barely affected by the occlusion colors, showing that the proposed model can handle heavily occluded facial regions with varying pixel values. 

\noindent\textbf{L2 vs. Adversarial Consistency:}
To show the effectiveness of our discriminator-based loss, we replace it with naive L2 loss, and the performance degrades, indicating that L2 loss is not enough to achieve consistency between clean and occluded/noisy images in terms of coefficients. The experiments are performed on three datasets: 1) \Real, 2) \Synthetic, and 3) noisy face set, as shown in Table~\ref{tab:Reax}, \ref{tab:synx_} and \ref{tab:noise}, respectively.

\begin{table}[!ht]

    \centering
  \resizebox{\columnwidth}{!} { \begin{tabular}{c c cccccccccc}
 \hline
   \multicolumn{3}{c}{} &&\multicolumn{7}{c}{Perceptual Error ($\downarrow$)}\\
   \cmidrule{5-11}
       \multicolumn{3}{c}{Losses} & &VGG-Face&FaceNet&FaceNet-$512$&OpenFace&DeepFace&ArcFace&SFace\\
         \midrule

         \multicolumn{3}{c}{L2}&  & $0.879\pm 0.132$& $1.135\pm 0.147$& $1.184\pm 0.149$&$0.751\pm 0.186$&$0.648\pm 0.185$&$1.283\pm 0.138$&$1.157\pm 0.162$\\
    \multicolumn{3}{c}{Adversarial} & &$\mathbf{0.786 \pm 0.138}$ &$\mathbf{0.943\pm 0.187}$&$\mathbf{0.918\pm 0.171}$&$\mathbf{0.682\pm 0.189}
        $&$\mathbf{0.603\pm 0.176}$&$\mathbf{1.025\pm 0.168}$&$\mathbf{1.052\pm 0.140}$\\ 

        \bottomrule
                           
    \end{tabular}}  \caption{A quantitative comparison of the perceptual error metric with other approaches on the proposed \Real~dataset, where the error numbers are the lower the better. }
		 
    \label{tab:Reax}
 
\end{table}

\begin{table}[!ht] 
    
    \centering
  \resizebox{\columnwidth}{!} { \begin{tabular}{c c cccccccccc}
    \hline
   \multicolumn{3}{c}{} &&\multicolumn{7}{c}{Perceptual Error ($\downarrow$)}\\
      \cmidrule{5-11}
        \multicolumn{3}{c}{Losses} & &VGG-Face&FaceNet&FaceNet-$512$&OpenFace&DeepFace&ArcFace&SFace\\
         \midrule

           \multicolumn{3}{c}{L2} &  & $0.895\pm 0.140$& $1.0753\pm 0.150$& $1.116\pm 0.141$&$0.880\pm 0.196$&$0.954\pm 0.159$&$1.227\pm 0.159$&$1.114\pm 0.152$\\

      \multicolumn{3}{c}{Adversarial}   &  &$\mathbf{0.802\pm 0.151}$&$\mathbf{0.891\pm 0.179}$&$\mathbf{0.879\pm 0.174}
        $&$\mathbf{0.871\pm 0.195}$&$\mathbf{0.847\pm 0.165}$&$\mathbf{0.983\pm 0.186}$&$\mathbf{1.016\pm 0.168}$\\ 

        \bottomrule
                           
    \end{tabular}}    \caption{A quantitative comparison of the perceptual error with other approaches on the proposed \Synthetic~dataset, where the error numbers are the lower the better. } 
\label{tab:synx_}
\end{table}
\begin{table}[!ht]   
    \centering
  \resizebox{\columnwidth}{!} { \begin{tabular}{c c cccccccccc}
    \hline
   \multicolumn{3}{c}{} &&\multicolumn{7}{c}{Perceptual Error ($\downarrow$)}\\
      \cmidrule{5-11}
       \multicolumn{3}{c}{Losses} & &VGG-Face&FaceNet&FaceNet-$512$&OpenFace&DeepFace&ArcFace&SFace\\
         \midrule

       \multicolumn{3}{c}{L2}  &  & $0.943\pm 0.157$&$1.197\pm 0.184$ & $1.198\pm 0.161$&$0.774\pm 0.185$&$0.676\pm 0.180$&$1.239\pm 0.116$&$1.238\pm 0.106$\\
        
        \multicolumn{3}{c}{Adversarial} &  &$\mathbf{0.798\pm 0.173}$&$\mathbf{0.976\pm 0.198}$&$\mathbf{0.959\pm 0.187}
        $&$\mathbf{0.706\pm 0.195}$&$\mathbf{0.611\pm 0.181}$&$\mathbf{1.014\pm 0.148}$&$\mathbf{1.127\pm 0.123}$\\ 

        \bottomrule
                           
    \end{tabular}} \caption{
		A quantitative comparison of the perceptual distance using perceptual error metric with other approaches on the proposed noisy variant of CelebA-test dataset, where the error numbers are the lower the better. } 
    \label{tab:noise}

\end{table}

\noindent\textbf{Choice of Weights for the Losses}: To choose the weights associated with various losses for training the \MainFramework, we perform several experiments as shown in Table~\ref{table:weight_losses} and \ref{table:real_life_weights_}. In Table~\ref{table:weight_losses}, we vary the weights $\beta_O$ and $\beta_N$ associated with occlusion- and noise-resistive photometric losses by fixing $\beta_C=0.001$ corresponding to consistency loss. Besides, to obtain the best performance of the model against the consistency loss, we fix $\beta_O=\beta_N=1.92$ and vary the weight $\beta_C$ in Table~\ref{table:real_life_weights_}. Based on the results in the tables, we conclude that the finest performance of the proposed model is obtained at $\beta_O=\beta_N=1.92$ and $\beta_C=0.001$. 	

\begin{table}[!ht]
\centering{\footnotesize{
\scriptsize{\resizebox{\columnwidth}{!}{\begin{tabular}{cc c c c c cc c c c ccccc}\toprule
&&&&&& &\multicolumn{9}{c}{Perceptual Error ($\downarrow$)} \\ \cmidrule{8-16}
\multicolumn{2}{c}{Weights}&&&& && Real&&&&Synthetic&&&& Noisy \\ \midrule
 			$0.48$ & $0.48$   &&&  & &   &    {$1.092\pm 0.135$} &&&& {$1.097\pm 0.162$}  &&&&  {$1.159\pm 0.122$}  \\ 		 

			$0.48$ & $1.92$  & &&&&       &   {$1.093\pm 0.135$}  &&&& {$1.097\pm 0.162$} &&&&   {$1.143\pm 0.122$}  \\ 

			$1.62$ &$1.62$    &&&&&     & {$1.071\pm 0.138$} &&&&   {$1.043\pm 0.166$}  &&&& {$1.135\pm 0.123$}   \\

 			$1.92$ & $1.92$   &&&&&      &  {$\mathbf{1.052\pm 0.140}$} &&&&   {$\mathbf{1.016\pm 0.168}$}  && && {$\mathbf{1.127\pm 0.123}$}    \\
 		    $2.22$ &  $2.22$  &&&  & &    &  {${1.097\pm 0.134}$} && &&  {${1.033\pm 0.167}$}  &&&& {${1.153\pm 0.122}$}    \\
 			$2.84$ & $2.84$  &&&  &   &  &  {$1.153\pm 0.164$}  &&&& {$1.45\pm 0.155$} &&&&  {$1.201\pm 0.118$}  \\\bottomrule
 		\end{tabular}}}}} \caption{Impact of weights associated with occlusion- and noise-resistive photometric losses on our results. For this purpose, we fix $\beta_C=10^{-3}$. The results are obtained using SFace.}\label{table:weight_losses}
\end{table}

\begin{table}[!ht]

\centering{\footnotesize{
\scriptsize{\resizebox{\columnwidth}{!}{\begin{tabular}{cccccccccccccc}\toprule
&&& &&&&\multicolumn{7}{c}{Perceptual Error ($\downarrow$)} \\ 
\cmidrule{8-14}
\multicolumn{1}{c}{Weights}&&& &&&& Real&&&Synthetic&&& Noisy \\ \midrule

	$0.0001$ &&&&&&&$0.148\pm 0.163$& &&$1.420\pm 0.0154$&&& $0.196\pm 0.118$ \\ 		 

	$0.001$ &&&& &&& {$\mathbf{1.052\pm 0.140}$} &&& {$\mathbf{1.016\pm 0.168}$}& && {$\mathbf{1.127\pm 0.123}$}\\

		$0.01$&& &&&&& $1.157\pm 0.161$ & && $1.125\pm 0.149$&&& $1.206\pm 0.117$ \\

 			\bottomrule
 		\end{tabular}}}}} 
			\caption{Impact of weights associated with discriminator loss on our results. For this purpose, we fix $\beta_O=\beta_N=1.92$. The results are obtained using SFace.}	\label{table:real_life_weights_}
 	
\end{table}

\noindent\textbf{Impact of Various Losses}: 
To study the impact of losses for the occlusions, we perform a quantitative analysis on the losses for the various cases: \textit{real-world occlusions}, \textit{synthetic occlusions} and \textit{noisy face} images. 
To answer this question, we train models with various combinations of the proposed losses and qualitatively evaluate them.
We show these quantitative results in 
Table~\ref{tab:Rea} and~\ref{tab:sync}, $\mathcal{L}_O$ is insufficient to tackle real-world and synthetic occlusions. In addition, $\mathcal{L}_N$ is not effective for addressing the issue of occlusions. The combination of $\mathcal{L}_O$ and $\mathcal{L}_N$ shows no significant improvement in performance. 
However, the combination of the consistency loss $\mathcal{L}_C$ and $\mathcal{L}_O$ significantly improves the $3$D vertex accuracy for delusional occlusions. Finally, by exploiting the losses $\mathcal{L}_O$, $\mathcal{L}_N$, and $\mathcal{L}_C$ altogether, we obtain the best performance of the proposed model. 
For the noisy image, Table~\ref{tab:nois} shows that the loss $\mathcal{L}_O$ does not contribute towards the improvement in the model performance. Besides, $\mathcal{L}_N$ alone is insufficient to tackle the face image noise. An improvement is observed in exploiting $\mathcal{L}_C$ for training the proposed model. The results demonstrate that the best model performance for noisy inputs can be obtained by either deploying a combination of $\mathcal{L}_C$ and $\mathcal{L}_N$ or using all three losses.
We conjecture that the loss $\mathcal{L}_O$ is
dedicated to addressing the issue of occlusions.
Thus, the improvement is not significant when all three losses are used as compared to the grouping of 
$\mathcal{L}_N$ and $\mathcal{L}_C$.
However, their cumulative usage is crucial for real-life scenarios as the face images contain both occlusions and noise.

\begin{table}[!ht]

    \centering
  \resizebox{\columnwidth}{!} { \begin{tabular}{c c cccccccccc}
 \hline
   \multicolumn{3}{c}{Losses} &&\multicolumn{7}{c}{Perceptual Error ($\downarrow$)}\\
    \cmidrule{1-3}  \cmidrule{5-11}
         $\mathcal{L}_O$& $\mathcal{L}_N$& $\mathcal{L}_C$& &VGG-Face&FaceNet&FaceNet-$512$&OpenFace&DeepFace&ArcFace&SFace\\
         \midrule
         &&&  & $0.952\pm 0.129$& $1.202\pm 0.146$& $1.237\pm 0.141$&$0.769\pm 0.189$&$0.656\pm 0.187$&$1.313\pm 0.114$&$1.215\pm 0.119$\\
         
         $\checkmark$&&&  & $0.874\pm 0.134$& $1.116\pm 0.153$& $1.177\pm 0.151$&$0.746\pm 0.189$&$0.641\pm 0.187$&$1.278\pm 0.129$&$1.151\pm 0.126$\\
         &$\checkmark$&&  & $0.952\pm 0.129$& $1.202\pm 0.146$& $1.237\pm 0.141$&$0.769\pm 0.189$&$0.656\pm 0.187$&$1.313\pm 0.114$&$1.215\pm 0.119$\\
         &&$\checkmark$&  & $0.813\pm 0.136$& $1.034\pm 0.178$& $1.015\pm 0.163$&$0.703\pm 0.189$&$0.624\pm 0.181$&$1.243\pm 0.159$&$1.097\pm 0.134$\\
         
         $\checkmark$&$\checkmark$&&  & $0.874\pm 0.134$& $1.116\pm 0.153$& $1.177\pm 0.151$&$0.746\pm 0.189$&$0.641\pm 0.187$&$1.278\pm 0.129$&$1.151\pm 0.126$\\
       $\checkmark$ && $\checkmark$& &$\mathbf{0.786 \pm 0.138}$ &$\mathbf{0.943\pm 0.187}$&$\mathbf{0.918\pm 0.171}$&$\mathbf{0.682\pm 0.189}
        $&$\mathbf{0.603\pm 0.176}$&$\mathbf{1.025\pm 0.168}$&$\mathbf{1.052\pm 0.140}$\\   
       &$\checkmark$&$\checkmark$&  & $0.813\pm 0.136$& $1.034\pm 0.178$& $1.015\pm 0.163$&$0.703\pm 0.189$&$0.624\pm 0.181$&$1.243\pm 0.159$&$1.097\pm 0.134$\\
       $\checkmark$ &$\checkmark$& $\checkmark$& &$\mathbf{0.786 \pm 0.138}$ &$\mathbf{0.943\pm 0.187}$&$\mathbf{0.918\pm 0.171}$&$\mathbf{0.682\pm 0.189}
        $&$\mathbf{0.603\pm 0.176}$&$\mathbf{1.025\pm 0.168}$&$\mathbf{1.052\pm 0.140}$\\ 

        \bottomrule
                           
    \end{tabular}} \caption{A quantitative comparison of the perceptual error with other approaches on the proposed \Real~dataset, where the error numbers are the lower the better. } 
		 
    \label{tab:Rea}
 
\end{table} 	

\begin{table}[!ht]    
    \centering
  \resizebox{\columnwidth}{!} { \begin{tabular}{c c cccccccccc}
    \hline
   \multicolumn{3}{c}{Losses} &&\multicolumn{7}{c}{Perceptual Error ($\downarrow$)}\\
    \cmidrule{1-3}  \cmidrule{5-11}
         $\mathcal{L}_O$& $\mathcal{L}_N$& $\mathcal{L}_C$& &VGG-Face&FaceNet&FaceNet-$512$&OpenFace&DeepFace&ArcFace&SFace\\
         \midrule
        &&&  & $0.952\pm 0.130$& $1.157\pm 0.133$& $1.195\pm 0.126$&$0.894\pm 0.196$&$0.854\pm 0.165$&$1.284\pm 0.150$&$1.241\pm 0.129$\\
        $\checkmark$&&&  & $0.894\pm 0.140$& $1.065\pm 0.152$& $1.056\pm 0.147$&$0.880\pm 0.196$&$0.851\pm 0.165$&$1.167\pm 0.164$&$1.143\pm 0.154$\\       
       &$\checkmark$&&  & $0.952\pm 0.130$& $1.157\pm 0.133$& $1.195\pm 0.126$&$0.894\pm 0.196$&$0.854\pm 0.165$&$1.284\pm 0.150$&$1.241\pm 0.129$\\       
        
 &&$\checkmark$&  & $0.823\pm 0.146$& $0.936\pm 0.161$& $0.978\pm 0.159$&$0.879\pm 0.196$&$0.849\pm 0.165$&$1.093\pm 0.179$&$1.098\pm 0.161$\\

        $\checkmark$&$\checkmark$&&  & $0.894\pm 0.140$& $1.065\pm 0.152$& $1.056\pm 0.147$&$0.880\pm 0.196$&$0.851\pm 0.165$&$1.167\pm 0.164$&$1.143\pm 0.154$\\  
    $\checkmark$   & & $\checkmark$ &  &$\mathbf{0.802\pm 0.151}$&$\mathbf{0.891\pm 0.179}$&$\mathbf{0.879\pm 0.174}
        $&$\mathbf{0.871\pm 0.195}$&$\mathbf{0.847\pm 0.165}$&$\mathbf{0.983\pm 0.186}$&$\mathbf{1.016\pm 0.168}$\\ 
       &$\checkmark$&$\checkmark$&  & $0.823\pm 0.146$& $0.936\pm 0.161$& $0.978\pm 0.159$&$0.879\pm 0.196$&$0.849\pm 0.165$&$1.093\pm 0.179$&$1.098\pm 0.161$\\       
 
    $\checkmark$   &$\checkmark$ &$\checkmark$  &  &$\mathbf{0.802\pm 0.151}$&$\mathbf{0.891\pm 0.179}$&$\mathbf{0.879\pm 0.174}
        $&$\mathbf{0.871\pm 0.195}$&$\mathbf{0.847\pm 0.165}$&$\mathbf{0.983\pm 0.186}$&$\mathbf{1.016\pm 0.168}$\\ 

        \bottomrule
                           
    \end{tabular}}
\caption{A quantitative comparison of the perceptual error with other approaches on the proposed \Synthetic~dataset, where the error numbers are the lower the better. }  
    \label{tab:sync}
\end{table}

\begin{table}[!ht]    
    \centering
  \resizebox{\columnwidth}{!} { \begin{tabular}{c c cccccccccc}
    \hline
   \multicolumn{3}{c}{Losses} &&\multicolumn{7}{c}{Perceptual Error ($\downarrow$)}\\
    \cmidrule{1-3}  \cmidrule{5-11}
         $\mathcal{L}_O$& $\mathcal{L}_N$& $\mathcal{L}_C$& &VGG-Face&FaceNet&FaceNet-$512$&OpenFace&DeepFace&ArcFace&SFace\\
         \midrule
       & & &  & $0.934\pm 0.151$& $1.209\pm 0.187$& $1.203\pm 0.159$&$0.799\pm 0.183$&$0.689\pm 0.178$&$1.283\pm 0.107$&$1.256\pm 0.115$\\    
       
         $\checkmark$& & &  & $0.934\pm 0.151$& $1.209\pm 0.187$& $1.203\pm 0.159$&$0.799\pm 0.183$&$0.689\pm 0.178$&$1.283\pm 0.107$&$1.256\pm 0.115$\\

       &   $\checkmark$ & &  & $0.883\pm 0.162$&$1.195\pm 0.188$ & $1.140\pm 0.175$&$0.768\pm 0.188$&$0.671\pm 0.180$&$1.191\pm 0.123$&$1.198\pm 0.118$\\
         
        & &  $\checkmark$&  & $0.809\pm 0.168$&$1.096\pm 0.194$ & $1.074\pm 0.183$&$0.741\pm 0.191$&$0.637\pm 0.180$&$1.085\pm 0.136$&$1.153\pm 0.121$\\
      $\checkmark$    &   $\checkmark$ & &  & $0.883\pm 0.162$&$1.195\pm 0.188$ & $1.140\pm 0.175$&$0.768\pm 0.188$&$0.671\pm 0.180$&$1.191\pm 1.123$&$1.198\pm 0.118$\\       
        $\checkmark$& &  $\checkmark$&  & $0.809\pm 0.168$&$1.096\pm 0.194$ & $1.074\pm 0.183$&$0.741\pm 0.191$&$0.637\pm 0.180$&$1.085\pm 0.136$&$1.153\pm 0.121$\\         

        &$\checkmark$ &$\checkmark$ &  &$\mathbf{0.798\pm 0.173}$&$\mathbf{0.976\pm 0.198}$&$\mathbf{0.959\pm 0.187}
        $&$\mathbf{0.706\pm 0.195}$&$\mathbf{0.611\pm 0.181}$&$\mathbf{1.014\pm 0.148}$&$\mathbf{1.127\pm 0.123}$\\ 
        $\checkmark$&$\checkmark$ &$\checkmark$ &  &$\mathbf{0.798\pm 0.173}$&$\mathbf{0.976\pm 0.198}$&$\mathbf{0.959\pm 0.187}
        $&$\mathbf{0.706\pm 0.195}$&$\mathbf{0.611\pm 0.181}$&$\mathbf{1.014\pm 0.148}$&$\mathbf{1.127\pm 0.123}$\\ 

        \bottomrule
                           
    \end{tabular}}  \caption{
		A quantitative comparison of the perceptual error metric with other approaches on the proposed noisy variant of CelebA-test dataset, where the error numbers are the lower the better. }
    \label{tab:nois}

\end{table}

\section{More Discussions}
\subsection{R-Net and MoFA on Our Variant Datasets}


One question may arise: \textit{What if R-Net and MoFA are also trained using our occluded and noisy images?} Unfortunately, the original R-Net~\cite{deng2019accurate} is unsuitable to get trained on the variant datasets mainly due to the \textit{unreliable skin masks}.
Occlusions and image noise would distort the estimated skin masks, and the model may adapt to the distortions as the facial features. 
To make the model suitable to work with our occluded and noisy images, we exploit \textit{clean skin masks} instead of regressing the pixel values of projected $3$D faces directly on input images as in the original R-Net framework. 
Note that R-Net relies on the estimated skin masks to tackle minor occlusions, whereas our method can tackle heavy occlusion and noise issues without the mask dependency and tweaking tricks as mentioned above, indicating a wider usage of our \MainShort~framework.
The results for R-Net, re-trained on our variant datasets with clean skin masks, are much worse than ours, indicating that the proposed \RobustPipeline~can obtain better robustness than the skin mask technique, as shown in the second row of Table~\ref{table:real-life_2} and~\ref{table:noise_2}.
Finally, MoFA~\cite{tewari2017mofa} is also not designed to address heavy occlusions and noise. We re-trained MoFA with our variant datasets, and the reconstruction errors are much larger than ours, as shown in the first row of Table~\ref{table:real-life_2} and~\ref{table:noise_2}.
All the results show that the superior performance of our method is not solely because of the variant data. Compared with MoFA and R-Net, our \textbf{\MainShort}~framework can exploit all the variant data well, obtaining a more robust model for $3$D face reconstruction.\\
\textbf{Limitations with DECA:} DECA~\cite{feng2021learning} reproduces occlusions on the $3$D faces instead of removing them. Therefore, re-training of DECA on our dataset holds minimal significance.
\begin{table}[!ht]
\centering
{\scalebox{1.1}{\scriptsize{
		\begin{tabular}{c c c c c c c c c c c c c c}

		\toprule
	&&	&&&	& & \multicolumn{7}{c}{ {MICC}}\\
		\cmidrule{8-14}
		  \multicolumn{1}{c}{ {Methods}}&&&&& &&  {Cooperative} &&& Indoor && & Outdoor              \\ \midrule

		{MoFA} &&&  && &&{$2.29\pm 0.57$} & && {$2.29\pm 0.54$}   
	&	&&{$2.37\pm 0.62$}  \\ 

			{R-Net + clean mask}&&&&& && {$1.96\pm 0.54$}& & & {$1.95\pm 0.51$} && & {$2.03\pm 0.58$} \\
	\textbf{{	ROGUE (Ours)}} &&&&&&&{$\mathbf{1.69\pm 0.51}$}& && {$\mathbf{1.69\pm 0.47}$}&&  &{$\mathbf{1.73\pm 0.56}$}\\ \bottomrule
		\end{tabular}}}} 
		\caption{A quantitative analysis on the \textit{Synthetic Occlusion} variant of the standard MICC dataset~\cite{Bagdanov:2011:FHF:2072572.2072597}.}
		
\label{table:real-life_2}
\end{table}	



		  

 	
\begin{table}[!ht]

\centering{\scalebox{1.1}{\scriptsize{
		\begin{tabular}{cccccccc c c c c c c}

		\toprule
		&&&	& & \multicolumn{9}{c}{ {MICC}}\\
		\cmidrule{8-14}
		  \multicolumn{1}{c}{ {Methods}}&& &&&&&  {Cooperative}& && Indoor && & Outdoor              \\ \midrule
				MoFA &&& && && {$2.49\pm 0.65$} &&& {$2.49\pm 0.62$}  &&&{$2.66\pm 0.69$} \\ 
			 R-Net + clean mask& &&&&&& {$2.09\pm 0.59$}&&& {$2.09\pm 0.57$} &&& {$2.21\pm 0.61$}\\
			\textbf{ROGUE (Ours)}&&& &&&&{$\mathbf{1.85\pm 0.53}$} &&&{$\mathbf{1.85\pm 0.50}$} &&& {$\mathbf{1.91\pm 0.58}$}\\ \bottomrule
		\end{tabular}}}}
		 
		\caption{A quantitative analysis on the \textit{Noisy} variant of the standard MICC dataset~\cite{Bagdanov:2011:FHF:2072572.2072597}.
}	
\label{table:noise_2}
\end{table}

\subsection{Comparison with High-Fidelity Face Reconstruction Methods}
Recent GAN-based methods~\cite{gecer2019ganfit,chen2019photo,lin2020towards,piao2021inverting,gecer2021fast} show good performance of reconstructing facial details for $3$D faces, such as wrinkles, pores, folds, etc.
These methods focus on producing detailed high-fidelity facial textures, whereas \MainShort~focuses on addressing the issues of large-scale occlusions and noise for $3$D face reconstruction without posing additional dependencies, which are different aspects. 
Moreover, the ideas of these high-fidelity reconstruction methods are complementary to our method and can be integrated into our \GuidePipeline~to improve the texture details (without costing the robustness performance because of the \RobustPipeline), which could be our future work.

\subsection{Potential Negative Societal Impact}
The proposed method estimates the closest possible $3$D face from the occluded and noisy face images. The technology is highly efficient and effective but is not $100\%$ perfect. This may lead to cumbersome situations, mainly when used with face recognition systems for identifying masked criminals at the crime scene. The situation might end up in capturing an innocent person. Along with the impact mentioned above, invasion of privacy is also a concern with such a technology. The estimation of $3$D facial data from occluded images might raise concern among those who do not want to reveal their identity in certain situations. The negative impacts may be compensated by boosting the model's accuracy by deploying multiple copies of the same face image occluded with different patterns enabling the model to learn the significance of the visible region for reconstructing $3$D face geometry and texture. 

\subsection{Limitations}
The proposed \MainFramework~network addresses the issues of occlusions and noise in monocular face images for reconstructing a $3$D face. However, the method requires pre-processing of the images before serving them as the input to~\GuidePipeline~and~\RobustPipeline. This increases the net time required for training the proposed model. Further, the automated pre-processing may fail for several reasons, for example, the inability of the face detection model to detect the face in the image, thus posing a challenge to the proposed approach. This issue may be addressed by deploying domain adaptation or generalization techniques in which the model learns the $3$D face without requiring pre-processed face data. 
In addition, our method currently addresses occlusion and noise issues, which cover most but not all the distortion types for faces. This issue can be addressed by extending the ROGUE framework with more distortion types, such as blur or image compression.
We plan to investigate these directions in our future work.

\end{document}